\documentclass[11pt]{article}

\usepackage[preprint]{acl}

\usepackage{times}
\usepackage{latexsym}

\usepackage[T1]{fontenc}
\usepackage[utf8]{inputenc}

\usepackage{microtype}

\usepackage{inconsolata}

\usepackage{graphicx}

\usepackage{amsmath}
\usepackage{amssymb}
\usepackage{mathtools}
\usepackage{amsthm}
\usepackage{booktabs}
\usepackage{multirow}
\usepackage{arydshln}
\usepackage[table]{xcolor}
\usepackage{lineno}
\usepackage{xcolor}
\usepackage[inline]{enumitem}
\usepackage{makecell}
\usepackage{tabularx}
\usepackage{array}
\usepackage{colortbl}

\title{Learning to Select, Not Relearn: Hard-Routed Mixtures of Reasoning LoRAs}

\author{
\textbf{Seyed Alireza Molavi\textsuperscript{1}
\thanks{Corresponding author: 
\href{mailto:seyed.alireza.molavi@hh.se}
{\texttt{seyed.alireza.molavi@hh.se}}}},
\textbf{Zhan Su\textsuperscript{1}},
\textbf{Yan Hu\textsuperscript{2}}, \\
\textbf{Peyman Sheikholharam Mashhadi\textsuperscript{1}},
\textbf{Stefan Byttner\textsuperscript{1}},
\textbf{Prayag Tiwari\textsuperscript{1}} \\
\\
\textsuperscript{1}Halmstad University, Halmstad, Sweden \\
\textsuperscript{2}The Chinese University of Hong Kong, Shenzhen, China \\
\texttt{\{seyed.alireza.molavi, prayag.tiwari\}@hh.se}
}

\begin{document}
	\maketitle
	\begin{abstract} %% 171 words
		Composing independently trained LoRA adapters into a single large language model is useful for multi-domain adaptation, especially when the original training data cannot be shared. A common approach is to use MoE-style routing over LoRA experts, but for frozen pretrained adapters, soft weighted combinations can change the unit-scale additive update under which each LoRA module was originally trained. We propose \textbf{Hard-Routed MoR-LoRA}, a two-stage framework for composing frozen reasoning LoRA experts through unit-scale hard selection. First, domain-specific LoRA adapters are trained independently using reinforcement learning from verifiable feedback to obtain reasoning experts. Then, all experts are frozen, reasoning traces are distilled from them, and only a lightweight shared router together with a small attention LoRA is trained for integration. The router selects exactly one expert per token using hard top-1 routing, while a straight-through estimator enables gradient-based training. Experiments across five benchmarks, multiple model scales, and additional model families show that Hard-Routed MoR-LoRA preserves expert behavior while requiring substantially fewer trainable parameters than soft-routing mixture baselines. Our analysis further shows that normalized soft mixtures often concentrate most routing mass on a single expert, suggesting that hard unit-scale routing provides a simple and efficient abstraction for frozen LoRA expert composition\footnote{Our code is available at \href{https://github.com/sar-molavi/hard-routed-mor-lora}{github.com/sar-molavi/hard-routed-mor-lora}}.
	\end{abstract}
	
	\section{Introduction}
	
	Large language models (LLMs) have demonstrated strong performance across diverse language understanding and reasoning tasks \cite{zhang2025entropy,zhang2025ratt,brown2020language}. However, adapting these models to specialized domains remains expensive because full fine-tuning requires substantial computation and access to training data. Parameter-efficient fine-tuning (PEFT) methods address this issue by modifying only a small subset of parameters \cite{houlsby2019parameter}. Among them, Low-Rank Adaptation (LoRA) \cite{hu2022lora} is widely used because it adds lightweight trainable matrices while keeping the backbone frozen.
	
	In practice, domain adaptation is often done independently, where different groups train LoRA adapters on their own datasets but cannot share the original data due to privacy or regulatory constraints. Instead, they may release only the trained LoRA modules. A natural goal is therefore to compose multiple independently trained LoRA experts into a single model that supports heterogeneous tasks. Mixture-of-Experts (MoE) architectures \cite{shazeer2017outrageously,fedus2022switch} provide a natural mechanism for such modular reuse by routing each token to selected experts. However, when the goal is to reuse frozen independently trained LoRA adapters, directly applying standard soft MoE routing introduces a scale mismatch. LoRA modules are trained assuming a unit additive update, whereas MoE combines experts using routing weights. Soft routing, therefore, blends expert updates and changes the effective function applied by each adapter.  For example, LoRA-Mixer \cite{li2025loramixer} addresses this by retraining the adapters with auxiliary losses, but this increases trainable parameters, requires more data, and can alter the knowledge encoded in the original experts.
	
	This challenge becomes more pronounced for instruction-tuned models with chain-of-thought (CoT) reasoning ability. Conventional fine-tuning encourages imitation of short answers and may weaken internal reasoning behavior in large models, while reinforcement learning from verifiable feedback (RLVF) can produce stronger reasoning experts by optimizing correctness rather than matching reference outputs \cite{shao2024deepseekmath,guo2025deepseekr1}. However, existing LoRA composition methods primarily focus on supervised multi-task fine-tuning and do not study the training and integration of experts in instruction-tuned models that exhibit CoT reasoning.
	
	In this work, we propose \textbf{Hard-Routed MoR-LoRA}, a framework that separates reasoning acquisition from expert selection. First, domain-specific LoRA adapters are trained independently using RLVF to obtain reasoning. Then, the experts are frozen and integrated using a hard-routed mixture that selects exactly one expert per token through a straight-through estimator (STE). We distill reasoning traces from the experts and train only a shared lightweight router and a small attention adaptation using a small amount of data (1000 samples per dataset). Hard routing preserves the original LoRA assumption by applying each expert with unit scale, enabling modular composition without retraining the experts.
	
	Experiments across five benchmarks and multiple model scales show that RLVF produces stronger reasoning experts for capable models, while conventional fine-tuning can degrade reasoning behavior in large instruction-tuned models. We further demonstrate that hard routing achieves comparable or better performance than soft-routing baselines in our setting while using substantially fewer trainable parameters and keeping the pretrained experts frozen. Analysis of routing probabilities reveals that normalized soft top-$k$ mixtures implicitly behave like near hard top-1 selection, indicating that hard routing captures the intended behavior more directly.
	
	Our contributions are summarized as follows:
	\begin{itemize}[noitemsep, topsep=0pt]
		\item We identify a scaling mismatch that arises when soft MoE routing is applied to independently trained frozen LoRA adapters, and we use hard unit-scale routing to preserve standalone expert application without retraining experts.
		\item We introduce a two-stage framework that trains reasoning LoRA experts using RLVF and integrates them through distillation while keeping experts frozen.
		\item We provide empirical evaluation across model scales and tasks showing improved performance and parameter efficiency.
	\end{itemize}
	
	\section{Related Work}
	\paragraph{Modular LoRA adaptation.}
	LoRA enables efficient fine-tuning by learning small low-rank updates while keeping the backbone model frozen. Although effective for adapting a model to a single task, independently trained LoRA modules are usually task-specific and difficult to reuse across different domains, especially when the original training data cannot be shared. There is a rich body of research have Several works, therefore, studiedy how to combine multiple pretrained adapters into one model \cite{page2023multi,muqeeth2024learning,ostapenko2024towards,wen2025marvel}. LoraHub \cite{huang2023lorahub} learns weights that merge a set of LoRA modules into a single composed adapter, while LoRA-LEGO \cite{zhao2024merging} decomposes adapters into smaller rank components and reconstructs a merged adapter for cross-task reuse. These methods demonstrate that pretrained adapters can be combined without retraining, but the resulting model applies the same merged adapter to every input and cannot dynamically select different experts at the token level.
	
	\paragraph{Mixture-of-Experts with LoRA.}
    MoE architectures \cite{shazeer2017outrageously,fedus2022switch,ostapenko2023case,su2024mixture} increase model capacity by routing tokens to a sparse subset of experts. Instead of merging adapters into one module, MoE-based approaches keep multiple experts and select among them during inference. Recent work applies this idea to parameter-efficient tuning by treating LoRA modules as experts. MixLoRA \cite{li2024mixlora} inserts multiple LoRA modules into feed-forward layers (FFN) with top-$k$ routing. MoLE \cite{wu2024mixture} learns hierarchical gating across layer-wise adapters, while SiRA \cite{zhu2024sira} and GOAT \cite{li2025goat} study alignment and sparse routing strategies. The closest to our work, LoRA-Mixer, integrates LoRA modules into attention projections and trains a lightweight router. However, soft routing produces a weighted sum of LoRA updates, whereas standalone LoRA training assumes unit-scale application. In LoRA-Mixer, this issue is addressed by further fine-tuning the experts with regularization losses, which increases the number of trainable parameters and introduces additional integration-stage computation.
	
	\paragraph{Instruction-tuned Reasoning and Verifiable Feedback.}
	Instruction tuning and alignment stages shape how language models produce and structure reasoning \cite{ouyang2022training}. RLVF improves reasoning by optimizing correctness using automatically checkable rewards instead of matching reference outputs \cite{shao2024deepseekmath,guo2025deepseekr1}. DeepSeek-R1 \cite{guo2025deepseekr1} shows that strong reasoning ability can emerge from this training paradigm, and Group Relative Policy Optimization (GRPO) \cite{shao2024deepseekmath} stabilizes training without requiring a learned value critic. In our setting, adapters are added to instruction-tuned base models that already exhibit CoT behavior. Preserving this behavior during training is important, and RLVF provides a natural way to train reasoning experts without overriding the aligned reasoning patterns of the base model.
	
	\paragraph{Hard routing and Straight-through Estimators.}
	Hard routing selects a single expert through a discrete decision, which is not directly differentiable. The STE enables gradient-based training by using the hard decision in the forward pass and a surrogate gradient in the backward pass \cite{bengio2013estimating,liu2023bridging}. In our approach, STE allows top-1 expert selection with unit scaling, preserving the original LoRA formulation while still allowing end-to-end training of the router.
	
	\section{Preliminaries}
	This section briefly introduces the components utilized in our method: Low-Rank Adaptation, Mixture-of-Experts, and Reinforcement Learning from Verifiable Feedback.
	
	\subsection{Low-Rank Adaptation}
	LoRA parameterizes weight updates as a low-rank decomposition:
	\begin{align}\label{eq:lora}
		W_{\text{updated}} = W_{\text{pretrained}} + AB
	\end{align}
	where $A \in \mathbb{R}^{n \times r}$ and $B \in \mathbb{R}^{r \times n}$ with $r \ll n$. During fine-tuning, only $A$ and $B$ are optimized while $W_{\text{pretrained}}$ remains frozen, substantially reducing memory and computational cost.
	
	\subsection{Mixture-of-Experts Routing}
	Given token representation $x$, a router produces logits $\mathbf{G}(x) \in \mathbb{R}^{K}$ and the routing probabilities are formulated as:
	\begin{align}
		\mathbb{P}_i(x) = \operatorname{softmax}(\mathbf{G}(x))_i
	\end{align}
	Standard MoE combines multiple experts using routing weights \cite{fedus2022switch}, typically activating the top-$k$ experts and scaling each expert output by its routing probability. The final output is computed as
	\begin{align}
		\text{MoE}(x)
		= \sum_{i=1}^{K}
		\mathbb{I}\!\left(i \in \text{top-}k(\mathbb{P}(x))\right)
		\cdot \mathbb{P}_{i}(x)
		\cdot E_i(x)
	\end{align}
	where $\mathbb{I}(\cdot)$ denotes the indicator function and $E_i(x)$ is the $i$-th expert's output.
	
	\subsection{Reinforcement Learning from Verifiable Feedback}\label{sec:RL}
	
	RLVF trains models using rewards derived from automatically verifiable outputs rather than human preference annotations \cite{shao2024deepseekmath,guo2025deepseekr1}. 
	Unlike supervised fine-tuning, the model is not constrained to follow a reference reasoning trace; only the final answer is evaluated by the verifier. 
	This allows the model to freely generate intermediate CoT reasoning steps, encouraging the emergence of structured reasoning behavior.
	
	We adopt GRPO \cite{shao2024deepseekmath}. For a prompt $x$, a group of reasoning trajectories $\{\tau_i\}_{i=1}^N$ is sampled with rewards $\{r_i\}_{i=1}^N$:
	\begin{align}
		\bar r &= \frac{1}{N} \sum_{i=1}^N r_i \\
		A_i &= r_i - \bar r 
	\end{align}
	where $A_i$ is the advantage for the $i$-th reasoning trace. The reward function is detailed in Section~\ref{sec:rlvf-reward} and Table~\ref{table:rlvf-reproducibility}.
	
	To improve sample efficiency, we use traces collected under the behavior policy $\pi_{\text{old}}$, which is a few checkpoints behind the target policy $\pi_{\theta}$. For each trajectory $\tau_i$, we compute a token-level importance ratio:
	\begin{align}
		w_{i,t}
		= \frac{\pi_\theta(y_{i,t} \mid x, y_{i,<t})}
		{\pi_{\text{old}}(y_{i,t} \mid x, y_{i,<t})}
	\end{align}
	
	The off-policy token-level GRPO objective becomes:
	{\footnotesize
		\begin{align}
			\begin{split}
				\mathcal{L}_{\text{off}}(\theta)
				&= - \mathbb{E}_{x} \Bigg[
				\frac{1}{N} \sum_{i=1}^N A_i
				\sum_{t=1}^{T_i} w_{i,t}
				\log \pi_\theta\!\left(y_{i,t} \mid x, y_{i,<t}\right)
				\Bigg]
			\end{split}
		\end{align}
	}

	\section{Methodology}
	
	We propose \textbf{Hard-Routed MoR-LoRA (Mixture of Reasoning LoRA)}, a plug-in framework that integrates multiple reasoning LoRA experts into a unified model while preserving the original LoRA assumption and maintaining sparse computation. The key idea is to decouple \emph{reasoning acquisition} from \emph{expert selection}. First, each LoRA adapter independently learns domain-specific reasoning behavior using RLVF \cite{shao2024deepseekmath,guo2025deepseekr1}. Then, the pretrained experts are combined through a hard-routed mixture in which only a lightweight router and a small attention adaptation are trained using supervised fine-tuning (SFT) on distilled reasoning traces.
	
\begin{figure*}[ht]
		\centering
		\includegraphics[width=\linewidth]{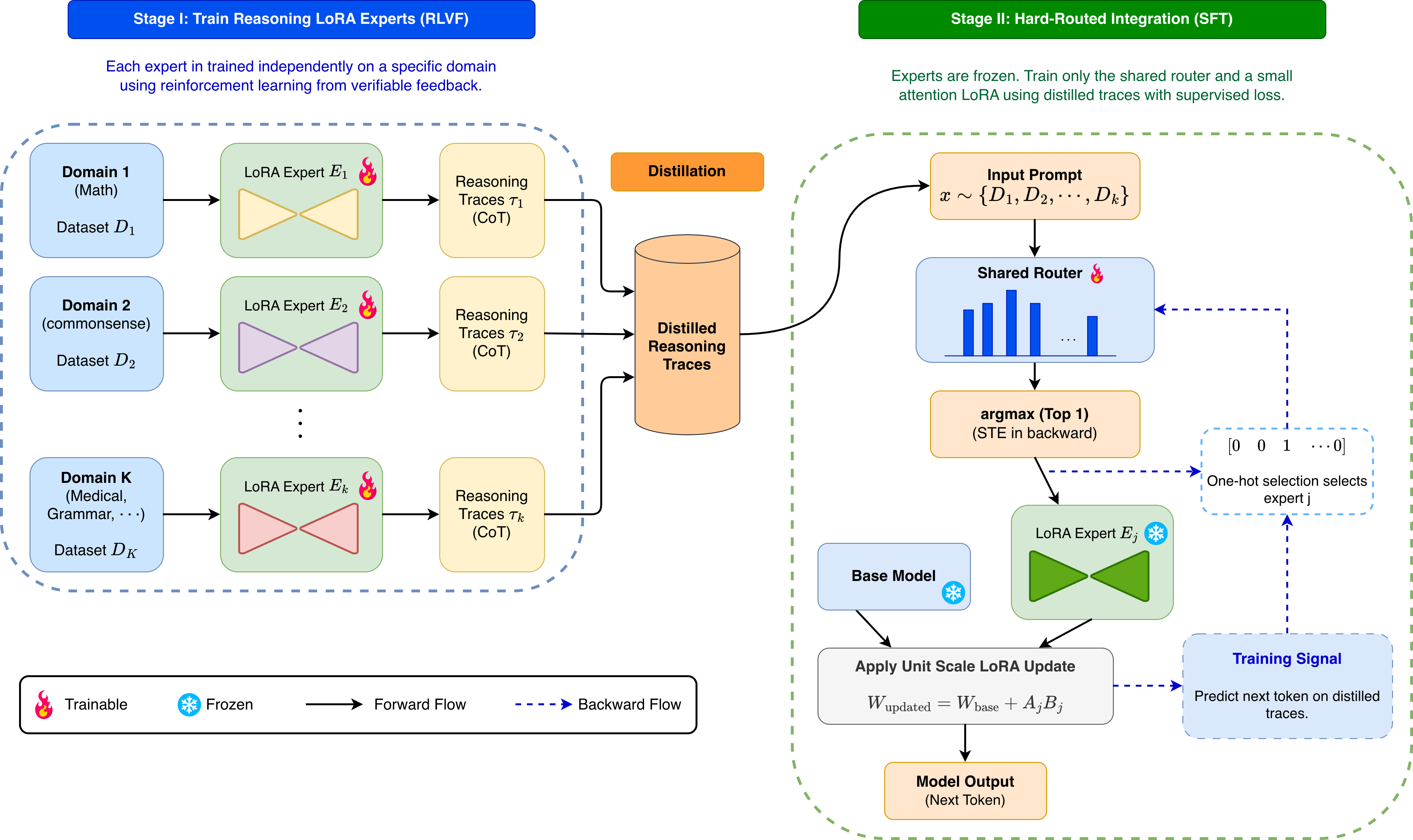}
		\caption{
			\textbf{Overview of the proposed Hard-Routed MoR-LoRA framework.}
			The method follows a two-stage training pipeline.
			\textbf{Stage~I (left):} Each LoRA adapter is independently trained on a specific domain using RLVF, producing a set of reasoning experts.
			\textbf{Stage~II (right):} All pretrained experts are integrated into a shared model and kept frozen. A shared lightweight router and a small attention LoRA are trained using supervised fine-tuning on distilled reasoning traces collected from the experts.
			During inference, the \textbf{router} selects a single expert per token, enforcing unit scaling and preserving the original LoRA assumption. STE enables gradients to propagate through the discrete routing decision by following the soft probabilities during the backward pass (dashed arrow).			
		}
		\label{fig:pipeline}
	\end{figure*}

	The overall two-stage pipeline is illustrated in Figure \ref{fig:pipeline}. Stage~I trains reasoning experts independently, while Stage~II integrates them into a shared model by learning which expert should be selected for each token.
	
	\subsection{Hard Mixture-of-LoRAs Architecture}
	
	\subsubsection*{Scale Mismatch in Soft LoRA Mixtures}
	From Eq.~\ref{eq:lora}, a pretrained LoRA expert is applied with unit contribution.
	However, the standard MoE applies weighted combinations of multiple experts:
	\begin{align}
		W_{\text{updated}} =
		W_{\text{pretrained}} +
		\sum_{i=1}^{k} \omega_i A_i B_i
	\end{align}
	where $\omega_i$ are routing probabilities, which change the magnitude of the update. Since each LoRA expert was trained assuming unit contribution, scaling by $\omega_i$ changes the effective adapter update used at integration time, creating a mismatch between standalone expert training and mixture usage. LoRA-Mixer compensates for this mismatch by further fine-tuning the adapters with preservation losses, aiming to limit deviation from the pretrained LoRA parameters.
	
	\subsubsection*{Hard Routing with Straight-Through Estimator}
	
	To preserve the original LoRA assumption, the selected expert must be applied with unit scale. However, a softmax router produces continuous weights. In a top-1 soft-routing variant, one may first select the largest routing probability $p_j$ and apply the selected expert as $p_j E_j(x)$. This remains differentiable with respect to the router logits, but it scales the selected LoRA update by $p_j < 1$, violating the unit-scale application under which the expert was trained. Alternatively, one can normalize over the selected top-1 set to enforce unit contribution, giving $p_j / p_j = 1$. However, this normalization is constant for any fixed selected expert and therefore provides zero gradient through the routing weight. The remaining dependence on the router is only through the discrete $\arg\max$ selection, which is non-differentiable. Therefore, a purely top-$1$ soft-routing formulation cannot simultaneously guarantee unit-scale expert application and provide a useful gradient for learning the router.
	
	Instead, we enforce \emph{top-1 routing} via discrete expert selection. For a token $x$:
	\begin{align}
		\mathbb{P} &= \operatorname{softmax}(\mathbf{G}(x)) \\
		j &= \arg\max_i( \mathbb{P}_{i})
	\end{align}
	
	This produces a one-hot routing decision. However, the $\arg\max$ operation is non-differentiable, preventing backpropagation through the router. 
	To enable training, we use the STE \cite{liu2023bridging}, which keeps the hard decision in the forward pass but approximates gradients using the soft probabilities in the backward pass:
	\begin{align}
		& Y_{\text{hard}} = \text{one-hot}(j) \\
		& Y = \mathrm{SG}(Y_{\text{hard}} - \mathbb{P}) + \mathbb{P}
	\end{align}
	where $\mathrm{SG}(\cdot)$ stops gradients. The forward pass uses the discrete routing vector $Y_{\text{hard}}$, while gradients propagate through $P$. The final output becomes
	\begin{equation}
		\mathrm{STE}(x)
		= \sum_{i=1}^K Y_i E_i(x)
		= E_j(x)
	\end{equation}
	which guarantees unit scaling of the selected LoRA expert while still allowing the router to be optimized via gradient descent.

	\paragraph{Parameter Efficiency.}
	All pretrained experts remain frozen. A single linear router is shared across layers, and only a small LoRA module on attention layers is trainable during integration. The number of trainable parameters, therefore, does not grow with the number of experts while maintaining sparse computation.
	
	\subsection{Training}\label{sec:training}
	There are two stages in the training: As shown in the Fig \ref{fig:pipeline}. In 
	\textbf{Stage~I:} we train domain-specific LoRA experts using RLVF \cite{shao2024deepseekmath,guo2025deepseekr1}. In 
	\textbf{Stage~II:} the LoRA adapters are frozen and we only train the router using SFT on distilled traces.
	
	\subsubsection*{Stage~I: Training Reasoning LoRA Experts}
	
	Each dataset trains one LoRA adapter independently. The adapters are optimized using RLVF with GRPO so that each expert specializes in domain-specific reasoning and produces CoT traces. After training, we obtain a set of reasoning experts:
	$
	\left\{ E_1, E_2, \dots, E_K \right\}
	$
	
	\subsubsection*{Stage~II: Mixture Integration via Distillation}
	
	\paragraph{Distillation Data Generation.}
	We query each trained expert with prompts from its domain and collect generated reasoning traces. %These traces form a supervised dataset representing expert behavior. 
	A limited number of samples is sufficient to construct the SFT dataset, since the objective of this Stage is to learn the routing policy rather than to relearn reasoning ability. The reasoning knowledge is already encoded in the frozen experts, and therefore, the training data only needs to cover representative input patterns for correct expert selection.

	\paragraph{Router and Attention Adaptation.}
	All experts remain frozen. We train \textbf{a shared linear router}, and \textbf{LoRA adapters on attention layers}
	using standard next-token likelihood on the distilled traces. The objective is to learn expert selection rather than relearn reasoning ability.

	\section{Experiments}
	
	We evaluate Hard-Routed MoR-LoRA along two independent dimensions:
	\begin{enumerate*}[label=(\arabic*)]
		\item the quality of reasoning experts obtained via RLVF, and
		\item the effectiveness of hard routing compared to soft mixture routing during expert integration.
	\end{enumerate*}
	
	\subsubsection*{Experimental Setup}
	
	We conduct experiments on instruction-following language model Meta-LLaMA-3 \cite{dubey2024llama} at three model scales, 1B, 3B, and 8B parameters. We also evaluate additional model families in Appendix~\ref{sec:additional}.
	
	\textbf{Datasets.} To evaluate performance across heterogeneous domains, we consider five representative tasks:
	\begin{enumerate*}[label=(\alph*)]
		\item mathematical reasoning,
		\item commonsense reasoning,
		\item medical multiple-choice question answering,
		\item reading comprehension, and
		\item grammatical understanding.
	\end{enumerate*}
	
	These tasks are evaluated using the GSM8K \cite{cobbe2021training}, ARC-Challenge (ARC-C) \cite{clark2018think}, Medical Question Answering (MedQA) \cite{jin2021disease}, BoolQ \cite{clark2019boolq}, and Corpus of Linguistic Acceptability (CoLA) \cite{warstadt2019neural} datasets, respectively. Each dataset covers a distinct domain with little to no overlap.

    \paragraph{Compared Models.}
    We use \textbf{Independent} to denote standalone dataset-specific LoRA adapters before mixture integration. These models are evaluated to measure the quality of individual experts, while the mixture methods evaluate how pretrained experts are integrated into one unified model.
    \begin{itemize}[noitemsep, topsep=-1pt]
    \item \textbf{Independent SFT:} one dataset-specific LoRA adapter trained with supervised fine-tuning on the FFN layers.
    \item \textbf{Independent RLVF:} one dataset-specific LoRA adapter trained with reinforcement learning from verifiable feedback on the FFN layers.
    \item \textbf{LoRAHub:} static integration by learning fixed merge weights over pretrained LoRA adapters.
    \item \textbf{LoRAMixer TopK=1:} soft top-1 routing with continued adapter training using an L2 preservation loss.
    \item \textbf{LoRAMixer TopK=2 Normalized:} normalized soft top-2 routing with continued adapter training using an L2 preservation loss.
    \item \textbf{MoLE:} MoE-style LoRA integration with soft routing and load-balancing regularization.
    \item \textbf{Ours (Hard-Routed MoR-LoRA):} hard top-1 routing via STE, with frozen experts and trainable router plus attention LoRA.
    \end{itemize}

    Unless otherwise stated, Stage~II integration uses at most 1000 samples per dataset. The independent adapters are trained separately for each dataset and then serve as the pretrained experts used by the integration methods.
	
	\subsection{RLVF Produces Stronger Reasoning Experts}\label{sec:rlvf-fn}
	
\begin{figure}[ht]
		\centering
		\includegraphics[width=\linewidth]{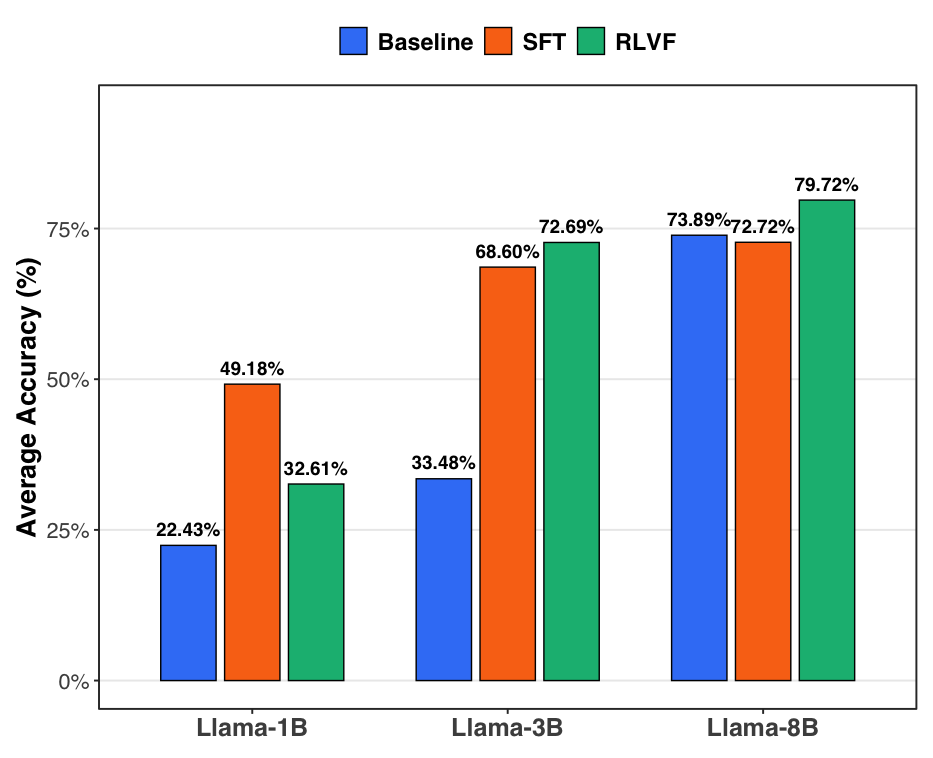}
		\caption{\textbf{Performance of standalone SFT and RLVF.} RLVF improves performance for 3B and 8B models but not for 1B, suggesting that reasoning-oriented training benefits models with sufficient capacity while preserving CoT behavior.}
		\label{fig:rlvf-fn}
	\end{figure}

	We first evaluate the quality of individual experts trained in Stage~I before any mixture integration. This isolates the effect of the training paradigm. Figure~\ref{fig:rlvf-fn} summarizes the average accuracy across model scales, while full per-task results are provided in Appendix~\ref{sec:app-1}, Table~\ref{table:fn-rlvf}.
	
	For medium and large models (3B and 8B), RLVF substantially improves performance on reasoning-intensive tasks. For example, the 3B model improves from 68.60 (SFT) to 72.69 with RLVF, and the 8B model improves from 72.72 to 79.72. Unlike fine-tuning, which encourages imitation of reference outputs, RLVF allows the model to explore intermediate reasoning steps and optimize correctness, leading to stronger multi-step reasoning behavior.
	
	The results also show that ordinary fine-tuning can degrade performance in large instruction-tuned models. We attribute this to fine-tuning, overriding the model's inherent COT behavior by encouraging short-form answer imitation. In contrast, RLVF preserves the aligned reasoning behavior because optimization depends only on verifiable correctness rather than matching a specific reasoning trace. For the 1B model, on the other hand, RLVF provides smaller gains compared to fine-tuning (32.61 vs 49.18). This is likely due to insufficient reasoning capacity as the 1B model struggles to produce diverse and correct reasoning trajectories, weakening the reward signal. Thus, the benefit of RLVF increases with model capability.
	
	As detailed in Table~\ref{table:fn-rlvf}, the improvement primarily appears in reasoning-heavy datasets such as \textsc{GSM8K} and \textsc{ARC-C}, while simpler linguistic tasks such as \textsc{CoLA} show limited advantage from RLVF. This supports the view that RLVF primarily strengthens reasoning behavior rather than general language modeling.
	
	\subsection{Hard Routing Preserves Frozen Expert Behavior During Integration}
	
\begin{figure*}[ht]
		\centering
		\includegraphics[width=\linewidth]{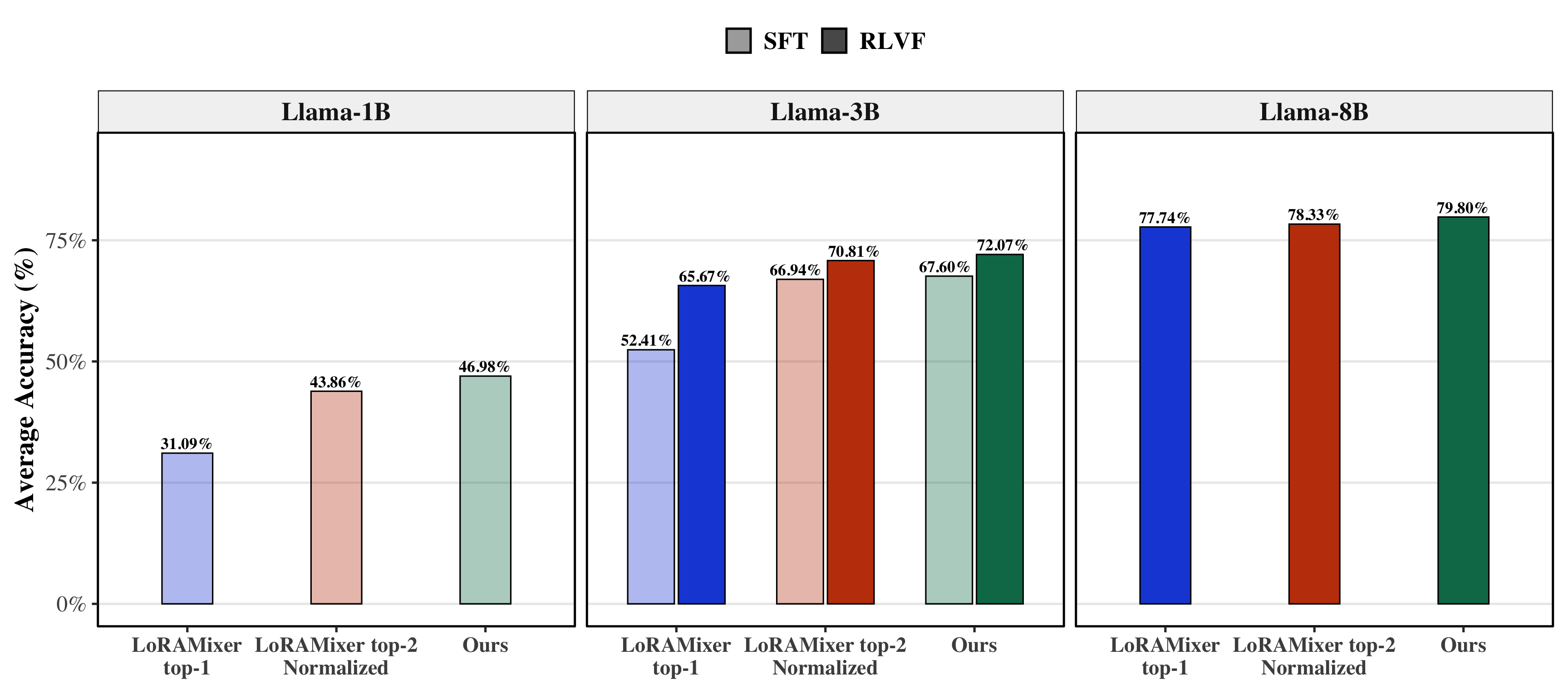}
		%\caption{\textbf{Hard routing vs. soft routing for LoRA composition.} Across model scales and expert types (SFT and RLVF), Hard-Routed MoR-LoRA achieves comparable or better performance than soft routing despite not retraining experts, indicating that routing quality dominates expert adaptation.}
        \caption{\textbf{Hard routing vs. soft routing for LoRA composition.} Fully saturated bars denote RLVF experts, whereas lighter bars denote SFT experts. Across model scales and expert types, Hard-Routed MoR-LoRA achieves comparable or better performance than soft routing despite not retraining experts, indicating that routing quality dominates expert adaptation.}
		\label{fig:mixer}
	\end{figure*}

	We now evaluate Stage~II, where pretrained experts are integrated using only 1000 samples per dataset. Figure~\ref{fig:mixer} summarizes the integration results for both SFT experts and RLVF experts, and Appendix~\ref{sec:app-2}, Tables~\ref{table:mixer-fn} and~\ref{table:mixer-rlvf} report the full numbers across five tasks and trainable parameter counts.
	
	In our frozen-expert integration setting, Hard-Routed MoR-LoRA achieves comparable or better performance than soft-routing baselines while requiring substantially fewer trainable parameters. Unlike LoRAMixer, which retrains the already-trained experts to compensate for the scaling mismatch introduced by soft routing, our architecture preserves the original LoRA formulation and therefore only learns expert selection while keeping the experts frozen. This suggests that much of the integration benefit comes from selecting appropriate frozen experts rather than extensively modifying their parameters. For example, on the 3B with RLVF experts, our method reaches 72.07 average while training $\approx$73M parameters, whereas LoRAMixer Top-1 reaches 65.67 while training $\approx$606M parameters.
	
	As shown in Figure~\ref{fig:mixer}, the improvement from hard routing over soft routing is larger for the 1B and 3B models compared to the 8B model. From Stage~I results, these smaller models benefit more from trained experts relative to the base model, so preserving expert behavior during integration is more important. Therefore, preserving the standalone behavior of each expert through hard unit-scale routing appears especially beneficial for smaller models. For the 8B model, the relative improvement is smaller but still consistent, with a particularly large gain on GSM8K as shown in Appendix Table~\ref{table:mixer-rlvf}, where the reasoning expert provides a strong improvement over the base model after RLVF training and hard routing better maintains its effectiveness.
	
	\subsection{Comparison with Additional LoRA Composition Baselines}
	
	The previous section compares Hard-Routed MoR-LoRA with LoRAMixer, which is the closest baseline to our setting. To provide a broader comparison, we also evaluate against two additional adapter-composition baselines: LoRAHub, a static adapter merging method, and MoLE, a representative MoE-LoRA method with load-balancing regularization. All methods are evaluated in the same Stage~II setting using RLVF-trained experts and at most 1000 integration samples per dataset.
	
	Table~\ref{table:main-additional-baselines} summarizes the results. LoRAHub performs static composition and therefore cannot dynamically select experts at inference time. LoRAMixer and MoLE use soft expert routing, whereas our method applies exactly one frozen LoRA expert with unit scale. Across LLaMA-3B and LLaMA-8B, Hard-Routed MoR-LoRA achieves the best average performance among the learned-composition baselines. The improvement over MoLE is larger on LLaMA-3B and smaller but still positive on LLaMA-8B. These results suggest that hard unit-scale routing is a simple and parameter-efficient integration strategy for frozen pretrained LoRA experts, avoiding expert retraining and additional load-balancing hyperparameters. The full per-task results are reported in Appendix \ref{sec:baselines}, Table \ref{table:baseline-rlvf-full}.
	
	\begin{table*}[ht]
		\centering
		\begin{tiny}
			\begin{sc}
				\begin{tabular}{l l c c c c}
					\toprule
					\textbf{Method} & \textbf{Composition Type} & \textbf{Routing Granularity} & \textbf{Extra Objective} & \textbf{LLaMA-3B Avg} & \textbf{LLaMA-8B Avg} \\
					\midrule
					LoRAHub & Static adapter merging & Prompt-level/static & Black-box weight search & 67.75 & 75.76 \\
					LoRAMixer TopK=1 & Soft mixture routing & Token-level & Preservation loss & 65.67 & 77.74 \\
					LoRAMixer TopK=2 Norm. & Normalized soft routing & Token-level & Preservation loss & 70.81 & 78.33 \\
					MoLE ($\alpha=0.5$) & Soft MoE-LoRA routing & Token-level & Load balancing & 68.50 & 78.50 \\
					MoLE ($\alpha=0.1$) & Soft MoE-LoRA routing & Token-level & Load balancing & 68.66 & 78.02 \\
                    Prompt Classification & None & Prompt-level/dynamic & BERT Classification & 72.79 & 79.87 \\ 
					\rowcolor{gray!10}
					\textbf{Hard-Routed MoR-LoRA} & \textbf{Hard unit-scale routing} & \textbf{Token-level} & \textbf{None} & \textbf{72.07} & \textbf{79.80} \\
					\bottomrule
				\end{tabular}
			\end{sc}
		\end{tiny}
		\caption{
			\textbf{Comparison with additional LoRA composition baselines using RLVF-trained experts.}
			All methods use the same Stage~II supervision budget of at most 1000 samples per dataset. Hard-Routed MoR-LoRA achieves the best average performance while preserving frozen expert behavior through hard unit-scale selection. The results also show that normalized soft routing remains competitive, but the hard routing provides a simpler and more parameter-efficient abstraction for frozen LoRA expert composition.
		}
		\label{table:main-additional-baselines}
	\end{table*}

	\subsection{Soft Routing Collapses Toward Single-Expert Selection}
	
	Hard routing resolves the LoRA integration mismatch by ensuring that exactly one expert with unit scaling is applied. In contrast, soft routing combines multiple experts using routing weights. While top-1 soft routing cannot maintain unit scaling, top-2 routing can be normalized so that the total update magnitude remains constant.
	%\begin{align}
	%	\begin{split}
		%		W_{\text{updated}}
		%		&= W_{\text{pretrained}}
		%		+ \omega_1 A_1 B_1 + \omega_2 A_2 B_2 \\
		%		&\text{where } \omega_1 + \omega_2 = 1
		%	\end{split}
	%\end{align}
	
	As shown in Table~\ref{table:mixer-fn} and Table~\ref{table:mixer-rlvf}, top-2 normalized routing performs substantially better than top-1 soft routing and approaches the performance of hard routing. However, examining the routing behavior reveals that the two experts are not used equally. 
	
\begin{figure}[ht]
		\centering
		\includegraphics[width=1\linewidth]{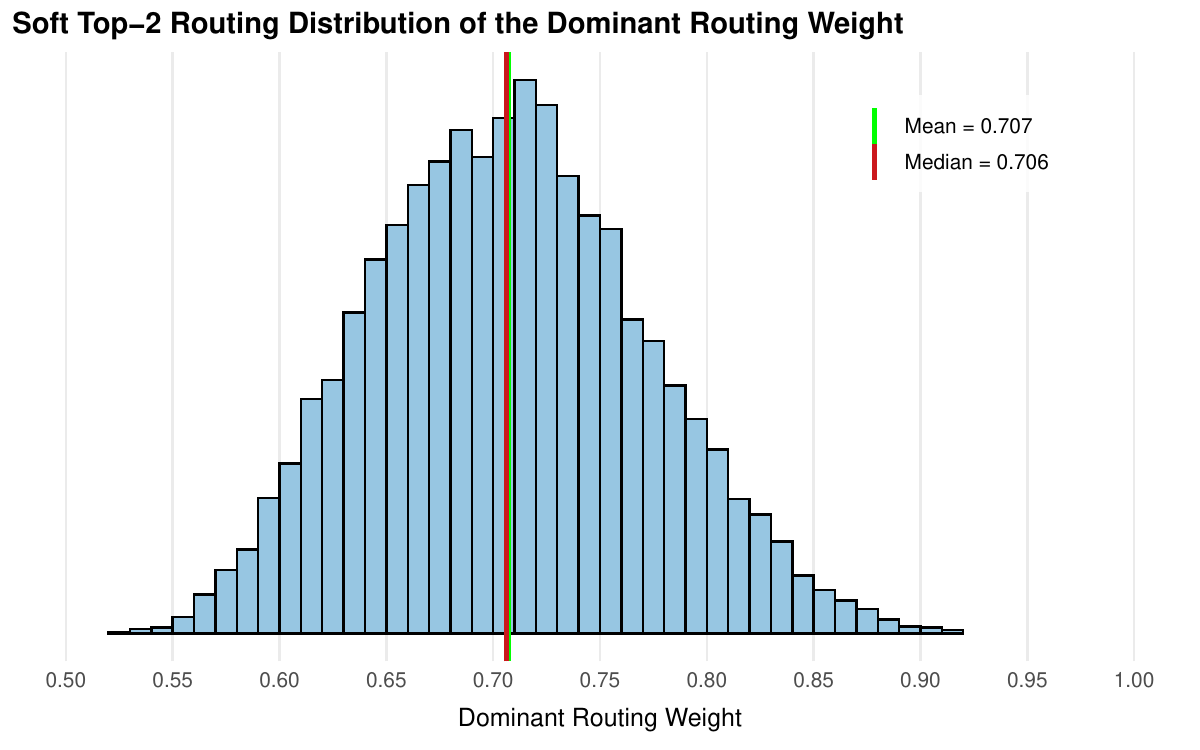}
		\caption{
			\textbf{Dominant routing weight under top-2 normalized routing.}
            The histogram is computed on the \textsc{GSM8K} using \texttt{Llama-3.2-3B-Instruct} with RLVF-trained experts. The dominant weight is the larger of the two selected expert weights after normalization. The average dominant weight is approximately $0.71$, and the average routing entropy is $0.6245$, indicating that soft top-2 routing often behaves like near single-expert selection.
		}
		\label{fig:routing}
	\end{figure}

	Figure~\ref{fig:routing} shows the distribution of the dominant routing weight, which is the larger weight among the two selected experts. The router usually assigns most of the probability mass to one expert, with an average dominant weight of about $0.71$ and an average routing entropy of $0.6245$. This shows that normalized top-2 routing often behaves like near top-1 routing. However, the model still evaluates and combines two LoRA experts, so the final update is a weighted mixture rather than the original unit-scale update of a standalone LoRA expert. Hard routing removes this extra step by directly selecting one expert, preserving sparsity and matching the original LoRA formulation more closely.
	
    \subsection{Ablation Study}
    We perform ablations to determine whether Stage~II improves performance through additional learning or mainly through better expert selection. We additionally analyze the Stage~II training budget in Appendix~\ref{sec:sample-budget}, the effect of attention LoRA rank in Appendix~\ref{sec:rank}, and alternative hard-routing approximations in Appendix~\ref{sec:hard-routing-approximations}.

	\subsubsection*{Is Reinforcement Learning Necessary for Integration?}
	
	Stage~II of our method aims to learn expert selection rather than relearning reasoning behavior. During this stage, all experts remain frozen, and only the shared router and attention LoRA are optimized. We compare two training strategies: (1) SFT on distilled reasoning traces generated by the experts, and (2) RLVF training for the routing selection.
	
	We present the results of the 3B model in Table~\ref{table:dist-rlvf}. RLVF achieves slightly higher performance (+0.93 average), but requires substantially higher computation due to trajectory sampling and advantage evaluation. The small improvement indicates that Stage~II primary learns which expert to select, while the reasoning capability is already encoded in the pretrained experts. Therefore, SFT on distilled traces is sufficient for integration, and reinforcement learning is primarily necessary only during expert training.
	
	\begin{table}[ht]
		\centering
		\begin{tiny}
			\begin{sc}
				\begin{tabular}{l cccccc}
					\toprule
					\textbf{Method} & GSM8K & ARC-C & MedQA & BoolQ & CoLA & Avg \\
					\midrule
					RLVF & 75.01 & 75.40 & 52.00 & 85.99 & 76.58 & \textbf{73.00} \\
					\rowcolor{gray!10}
					SFT & 73.62 & 75.43 & 51.53 & 83.09 & 76.70 & 72.07 \\
					\bottomrule
				\end{tabular}
				\caption{Comparison of training strategies for Stage~II integration on the 3B model. Both methods train only the router and attention LoRA while keeping experts frozen and using the same prompts.}
				\label{table:dist-rlvf}
			\end{sc}
		\end{tiny}
	\end{table}
	
	\subsubsection*{Prompt-Level and Mixed-Domain Routing}
    The original evaluation uses datasets with clearly separated domains. This raises a natural question: whether token-level routing is necessary, or whether the problem can mostly be solved by selecting one expert for the whole prompt. To test this, we add a prompt-level routing baseline. This baseline trains a \texttt{bert-base-uncased} \cite{devlin2019bert} classifier using the same Stage~II supervision budget, with at most 1000 examples per dataset. The classifier predicts the input domain and then applies the corresponding frozen LoRA expert to the entire prompt.

    The prompt-level classifier reaches 99.70\% domain classification accuracy. On LLaMA-3B with RLVF-trained experts, As shown in Table~\ref{table:main-additional-baselines}, prompt-level routing marginally outperforms Hard-Routed MoR-LoRA. This shows that prompt-level routing is a strong and simple baseline when each prompt belongs clearly to one domain. Thus, token-level routing should not be interpreted as necessary for clean single-domain inputs. Full per-task results and classifier details are provided in Appendix~\ref{sec:prompt-routing-details}.

    However, prompt-level routing has a structural limitation: it must select a single expert for the whole input. To test this limitation, we construct a mixed-domain evaluation where each prompt contains one GSM8K math problem and one BoolQ question, and the model is asked to produce two structured answers. No method is trained on mixed-domain prompts.

    \begin{table}[ht]
        \centering
        \begin{tiny}
        \begin{sc}
        \begin{tabular}{l ccc}
            \toprule
            Method & Math part & BoolQ part & Avg \\
            \midrule
            Prompt-level routing & \textbf{61.61} & 63.88 & 62.75 \\
            \rowcolor{gray!10}
            Hard-Routed MoR-LoRA & 59.04 & \textbf{73.00} & \textbf{66.02} \\
            \bottomrule
        \end{tabular}
        \end{sc}
        \end{tiny}
        \caption{
        \textbf{Mixed-domain evaluation with GSM8K and BoolQ.}
        Each input contains one math problem and one BoolQ question. No method is trained on mixed-domain prompts.
        }
        \label{table:mixed-domain}
    \end{table}

    As shown in Table~\ref{table:mixed-domain}, prompt-level routing performs better on the math part but considerably worse on the BoolQ part. This is expected because it must choose either the GSM8K expert or the BoolQ expert for the entire input. In contrast, ours is not restricted to one expert for the whole prompt and achieves better average performance. This result does not show that token-level routing fully solves mixed-domain reasoning, but it demonstrates why token-level expert selection is more flexible when a single input requires multiple expert behaviors. We additionally report unseen-dataset results on SVAMP and SST-2 in Appendix~\ref{sec:ood-eval}.

	%%%%%%%%%%%%%%%%%%%%%%%%%%%%%%%%%%%%%%

	\section{Conclusion}
	
	We study how independently trained LoRA adapters can be combined in instruction-tuned LLMs without changing their behavior. We find that a key challenge in frozen pretrained LoRA composition is the mismatch between unit-scale LoRA updates and weighted mixture routing. Prior LoRA-mixing methods often address this by adapting experts during integration, increasing trainable parameters. Hard-Routed MoR-LoRA instead treats integration as a routing problem: reasoning experts are first trained independently using RLVF, and the mixture is then trained only to select among them while keeping all experts frozen. Across multiple model scales and tasks, we observe three consistent results. First, verifiable-feedback training produces stronger reasoning experts for capable instruction-tuned models. Second, hard routing integrates experts more reliably and with far fewer trainable parameters than soft routing approaches. Third, normalized soft mixtures often concentrate most routing mass on one expert, explaining why hard unit-scale selection can be effective without retraining the experts. Overall, our results show that modular adaptation can be achieved by learning when to use each expert rather than modifying what each expert knows, enabling practical reuse under limited data or data-sharing constraints.

	\section{Limitations}
	Although our approach avoids retraining experts, integration is not completely training-free. Stage~II still requires a some amount of labeled data to learn the routing behavior. As a result, adding new experts or domains requires additional routing data and training, and the framework does not yet support true zero-shot integration. This limitation is more significant in settings where only very few or no samples are available.
	
	Our method is primarily evaluated in settings where inputs are associated with a small number of expert domains. In the clean single-domain setting, prompt-level routing is already highly competitive, showing that selecting one expert for the whole prompt can be sufficient when the domain is clear. At the same time, our mixed-domain GSM8K+BoolQ evaluation suggests a promising advantage of token-level routing: it is not restricted to a single expert decision for the entire input. With explicit mixed-domain training, the router could further learn when to switch experts within a prompt, making this formulation naturally suited to inputs that combine multiple types of reasoning. Similarly, our unseen-dataset results in Appendix~\ref{sec:ood-eval} suggest that the router can transfer useful expert behavior to related task distributions. A natural extension is to add an abstention or backbone route, allowing the model to bypass all LoRA experts and follow the original computation graph when none of the available experts is appropriate. These directions point toward more flexible expert composition with larger expert pools, mixed-domain supervision, and adaptive routing between specialized experts and the base model.

    \section*{Acknowledgment}
    This work was supported by the Carl Bennet AB PhD Program. Computational resources were provided by the National Academic Infrastructure for Supercomputing in Sweden (NAISS), funded by the Swedish Research Council.
	
	%\clearpage
	\bibliography{custom}
	
	\clearpage
	\appendix
	
	\section{Detailed Experimental Results}
	
	This section provides the \textbf{full per-task} results corresponding to the averaged metrics reported in the paper. All experiments follow the two-stage pipeline: training standalone experts (Stage~I) and integrating them through routing (Stage~II).
	
	\subsection{Stage~I: Standalone Expert Quality}\label{sec:app-1}
	
	Table~\ref{table:fn-rlvf} shows the performance of experts before mixture integration.
	We compare the original instruction-tuned model (Baseline), conventional fine-tuning (SFT), and RLVF.
	
	The detailed numbers match the observations in Section~\ref{sec:rlvf-fn}. RLVF improves reasoning-heavy tasks for 3B and 8B models, while fine-tuning can reduce performance in larger instruction-tuned models. By contrast, for the 1B model, RLVF gives smaller gains due to limited reasoning capacity.
	
	\subsection{Stage~II: Expert Integration}\label{sec:app-2}
	
	Tables~\ref{table:mixer-fn} and~\ref{table:mixer-rlvf} report mixture integration results for both fine-tuned experts and RLVF-trained experts.
	
	We compare LoRAMixer top-1 routing, LoRAMixer normalized top-2 routing, and ours (\textbf{Hard-Routed MoR-LoRA}). Across both expert types, hard routing achieves better performance while training far fewer parameters. This indicates that integration mainly depends on selecting the correct expert rather than modifying expert weights.
	
	\subsection*{Summary}
	
	Across all tables, three patterns are consistent:
	\begin{itemize}[nosep]
		\item RLVF produces stronger experts for capable models.
		\item Soft routing requires retraining experts and more parameters.
		\item Hard routing preserves expert behavior with fewer trainable parameters.
	\end{itemize}
	
	These tables confirm that the improvements in the paper are consistent across datasets and model sizes.
	
	\begin{table*}[!h]
		\centering
		\begin{small}
			\begin{sc}
				\begin{tabular}{l  c c c c c c}
					\toprule
					Method & GSM8K & ARC-C & MedQA & BoolQ & CoLA & Avg \\
					\midrule
					\midrule
					\multicolumn{6}{l}{\texttt{Llama-1B}} \\
					\midrule
					Baseline   & $03.49$  & $22.78$ & $26.71$ & $30.31$ & $28.86$ & $22.43$  \\
					\rowcolor{gray!5}
					SFT & $\mathbf{30.10}$ & $\mathbf{46.07}$ & $\mathbf{38.17}$ & $61.93$ & $\mathbf{69.61}$ & \underline{$\mathbf{49.18}$} \\
					\rowcolor{gray!15}
					RLVF & $00.05$ & $30.12$ & $34.33$ & $\mathbf{68.23}$ & $30.30$ & $32.61$ \\
					\midrule
					\midrule
					\multicolumn{6}{l}{\texttt{Llama-3B}} \\
					\midrule
					Baseline   & $01.82$  & $22.18$ & $33.70$ & $48.93$ & $60.79$ & $33.48$ \\
					\rowcolor{gray!5}
					SFT & $60.96$ & $69.97$ & $50.14$ & $83.43$ & $\mathbf{79.00}$ & $68.60$ \\
					\rowcolor{gray!15}
					RLVF & $\mathbf{75.36}$ & $\mathbf{76.88}$ & $\mathbf{51.13}$ & $\mathbf{84.43}$ & $75.65$ & \underline{$\mathbf{72.69}$} \\
					\midrule
					\midrule
					\multicolumn{6}{l}{\texttt{Llama-8B}} \\
					\midrule
					Baseline & $63.84$ & $81.38$ & $64.65$ & $82.53$ & $76.99$ & $73.89$ \\
					\rowcolor{gray!5}
					SFT & $70.74$ & $79.52$ & $60.10$ & $73.15$ & $\mathbf{80.09}$ & $72.72$ \\
					\rowcolor{gray!15}
					RLVF & $\mathbf{83.02}$ &  $\mathbf{82.94}$ & $\mathbf{66.06}$ & $\mathbf{87.68}$ & $78.91$ & \underline{$\mathbf{79.72}$} \\
					\bottomrule
				\end{tabular}
				\caption{
					\textbf{Performance of SFT and RLVF.} The Baseline row corresponds to the original model. RLVF improves reasoning performance for capable models (3B and 8B) while preserving chain-of-thought behavior, whereas fine-tuning can degrade CoT ability in large instruction-tuned models by overriding existing reasoning patterns. Smaller models (1B) benefit less from RLVF due to limited reasoning trajectory diversity.
				}
				\label{table:fn-rlvf}
			\end{sc}
		\end{small}
	\end{table*}
	
	\begin{table*}[!h]
		\begin{small}
			\begin{sc}
				\centering
				\begin{tabular}{l c c c c c c c}
					\toprule
					Method & GSM8K & ARC-C & MedQA & BoolQ & CoLA & Average & \makecell{\# Trainable\\Parameters} \\
					\midrule
					\midrule
					\multicolumn{8}{l}{\texttt{Llama-1B}} \\
					\midrule
					LoRAMixer TopK=1 &  $30.93$ & $56.76$ & $25.37$ & $39.72$ & $02.68$ & $31.09$ & $\approx 315M$\\
					\rowcolor{gray!5}
					\makecell[l]{LoRAMixer TopK=2\\Normalized}   &    $33.89$ & $45.82$ & $30.95$ &  $43.55$ & $65.10$ & $43.86$ & $\approx 315M$ \\
					\rowcolor{gray!15}
					\textbf{Ours} & $34.72$ & $45.05$ & $33.31$ & $53.18$ & $68.65$ & $ \mathbf{46.98}$ & $\approx 27M$ \\
					\midrule
					\midrule
					\multicolumn{8}{l}{\texttt{Llama-3B}} \\
					\midrule
					LoRAMixer TopK=1   &    $22.74$ & $46.50$ & $46.50$ &  $75.75$ & $70.57$ & $52.41$ & $\approx 606M$ \\
					\rowcolor{gray!5}
					\makecell[l]{LoRAMixer TopK=2\\Normalized}   &    $63.00$ & $66.13$ & $50.16$ &  $80.65$ & $74.78$ & $66.94$ & $\approx 606M$ \\
					\rowcolor{gray!15}
					\textbf{Ours} & $62.85$ & $66.98$ & $49.96$ & $82.00$ & $76.22$ & $ \mathbf{67.60}$ & $\approx 73M$ \\
					\bottomrule
				\end{tabular}
				\caption{\textbf{Stage~II. Integration using fine-tuning experts.} LoRAMixer retrains experts and requires substantially more trainable parameters, while Hard-Routed MoR-LoRA freezes experts and learns only routing and attention adaptation. Despite using fewer parameters, hard routing achieves comparable or better performance, showing that correct expert selection is more important than modifying expert weights.}
				\label{table:mixer-fn}
			\end{sc}
		\end{small}
	\end{table*}
	
	\begin{table*}[!h]
		\begin{center}
			\begin{small}
				\begin{sc}
					\begin{tabular}{l ccccccc}
						\toprule
						Method & GSM8K & ARC-C & MedQA & BoolQ & CoLA & Average &  \makecell{\# Trainable\\Parameters} \\
						\midrule
						\midrule
						\multicolumn{8}{l}{\texttt{Llama-3B}} \\
						\midrule
						LoRAMixer TopK=1 &  $63.53$ & $68.34$ & $52.40$ & $76.67$ & $67.40$ & $65.67$ & $\approx 606M$ \\
						\rowcolor{gray!5}
						\makecell[l]{LoRAMixer TopK=2\\Normalized} &  $69.67$ & $73.98$ & $52.08$ & $82.87$ & $75.46$ & $70.81$ & $\approx 606M$
						\\
						\rowcolor{gray!10}
						\textbf{Ours}       & $73.62$        & $75.43$        &  $51.53$       & $83.09$ & $76.70$ & $\mathbf{72.07}$ & $\approx 73M$   \\
						\midrule
						\midrule
						\multicolumn{8}{l}{\texttt{Llama-8B}} \\
						\midrule
						LoRAMixer TopK=1   & $79.51$     &  $83.30$       & $65.22$     & $83.70$ & $76.99$   & $77.74$ & $\approx 1.133B$ \\
						\rowcolor{gray!5}
						\makecell[l]{LoRAMixer TopK=2\\Normalized}   & $82.15$     &  $83.96$       & $66.06$     & $82.40$ & $77.08$   & $78.33$ & $\approx 1.133B$ \\
						\rowcolor{gray!10}
						\textbf{Ours}      & $84.69$        & $84.61$        & $67.17$        & $84.43$  & $78.09$  & $\mathbf{79.80}$ & $\approx 109M$   \\
						\bottomrule
					\end{tabular}
					\caption{\textbf{Stage~II. Integration using RLVF-trained reasoning experts.} Hard-Routed MoR-LoRA preserves expert behavior while requiring far fewer trainable parameters, outperforming soft routing methods that retrain experts and modify their learned reasoning patterns.}
					\label{table:mixer-rlvf}
				\end{sc}
			\end{small}
		\end{center}
	\end{table*}

    \subsection{Full Per-Task Results for Additional Baselines}\label{sec:baselines}

    In this section, we provide the full per-task results for the additional LoRA composition baselines. The paper reports average performance to keep the comparison compact, while the tables below show how each method behaves across individual datasets. We include three types of baselines: LoRAHub as a static adapter-merging method, LoRAMixer as a preservation-based soft-routing method, and MoLE as a MoE-LoRA method with load-balancing regularization. This allows us to compare static composition, soft expert mixing, and hard unit-scale routing under the same frozen-expert integration setting.

    All results in this section use RLVF-trained experts from Stage~I. During Stage~II, each method is trained with the same supervision budget of at most 1000 samples per dataset. For LoRAMixer, the pretrained adapters are further optimized with a preservation loss. For MoLE, we evaluate two load-balancing coefficients, $\alpha=0.5$ and $\alpha=0.1$, to account for sensitivity to the balancing objective. In contrast, Hard-Routed MoR-LoRA keeps all pretrained experts frozen and trains only the shared router and lightweight attention LoRA.

    Table~\ref{table:baseline-rlvf-full} provides the full per-task results. These results correspond to the averaged scores reported in Table~\ref{table:main-additional-baselines}.
    
    \begin{table*}[!h]
    	\centering
    	\begin{small}
    		\begin{sc}
    			\begin{tabular}{l c c c c c c}
    				\toprule
    				\textbf{Method} & GSM8K & ARC-C & MedQA & BoolQ & CoLA & Avg \\
    				\midrule
    				\midrule
    				\multicolumn{7}{l}{\texttt{LLaMA-3B}} \\
    				\midrule
    				LoRAHub & 72.55 & 68.86 & 51.06 & 74.20 & 72.10 & 67.75 \\
    				LoRAMixer TopK=1 & 63.53 & 68.34 & 52.40 & 76.67 & 67.40 & 65.67 \\
    				\makecell[l]{LoRAMixer TopK=2\\Normalized} & 69.67 & 73.98 & 52.08 & 82.87 & 75.46 & 70.81 \\
    				MoLE ($\alpha=0.5$) & 69.22 & 66.04 & \textbf{54.44} & 79.85 & 72.96 & 68.50 \\
    				MoLE ($\alpha=0.1$) & 69.37 & 65.70 & \textbf{54.67} & 79.76 & 73.83 & 68.66 \\
    				\rowcolor{gray!10}
    				\textbf{Hard-Routed MoR-LoRA} & \textbf{73.62} & \textbf{75.43} & 51.53 & \textbf{83.09} & \textbf{76.70} & \textbf{72.07} \\
    				\midrule
    				\midrule
    				\multicolumn{7}{l}{\texttt{LLaMA-8B}} \\
    				\midrule
    				LoRAHub & 82.41 & 84.64 & 65.51 & 76.21 & 70.04 & 75.76 \\
    				LoRAMixer TopK=1 & 79.51 & 83.30 & 65.22 & 83.70 & 76.99 & 77.74 \\
    				\makecell[l]{LoRAMixer TopK=2\\Normalized} & 82.15 & 83.96 & 66.06 & 82.40 & 77.08 & 78.33 \\
    				MoLE ($\alpha=0.5$) & 84.00 & \textbf{85.07} & 66.30 & 80.52 & 76.61 & 78.50 \\
    				MoLE ($\alpha=0.1$) & 83.93 & 84.90 & 64.96 & 79.91 & 76.41 & 78.02 \\
    				\rowcolor{gray!10}
    				\textbf{Hard-Routed MoR-LoRA} & \textbf{84.69} & 84.61 & \textbf{67.17} & \textbf{84.43} & \textbf{78.09} & \textbf{79.80} \\
    				\bottomrule
    			\end{tabular}
    		\end{sc}
    	\end{small}
    	\caption{
    		\textbf{Full per-task baseline comparison with RLVF-trained experts.}
    		Results are reported for LLaMA-3B and LLaMA-8B. Hard-Routed MoR-LoRA achieves the best average performance for both model sizes and outperforms the other composition baselines on most tasks.
    	}
    	\label{table:baseline-rlvf-full}
    \end{table*}

    \section{Sample Training Budget}\label{sec:sample-budget}
	We analyze how many supervision samples are required for Stage~II. For each dataset, we randomly sample prompts and generate distilled reasoning traces from the pretrained experts. The mixture model is then trained for one epoch on the combined distilled data from all five domains. We compare using 750 and 1000 prompts per dataset against using all available samples (30,573).
	
	Table~\ref{table:training-samples} shows that using only 1000 prompts per dataset achieves performance close to training on the full dataset, with less than $1\%$ difference in average accuracy. Increasing the amount of data yields only marginal gains, indicating that Stage~II mainly learns routing patterns rather than domain knowledge. Since reasoning ability is already encoded in the experts trained in Stage~I, only a small number of examples is sufficient to identify the appropriate expert.
	\begin{table}[!h]
		\centering
		\begin{tiny}
			\begin{sc}
				\begin{tabular}{c cccccc}
					\toprule
					\textbf{\makecell[c]{\# Samples}} & GSM8K & ARC-C & MedQA & BoolQ & CoLA & Avg \\
					\midrule
					$5 \times 750$ & 70.36 & 73.72 & 52.57 & 79.88 & 74.45 & 70.20 \\
					\rowcolor{gray!5}
					$5 \times 1000$ & 73.62 & 75.43 & 51.53 & 83.09 & 76.70 & 72.07 \\
					\rowcolor{gray!15}
					\textbf{\makecell[c]{ALL\\(30,573)}} & 75.32 & 75.02 & 52.32 & 85.12 & 77.10 & \textbf{72.98} \\
					\bottomrule
				\end{tabular}
				\caption{Effect of the Stage~II training budget on the 3B model. The first two rows use the specified number of prompts per dataset (five datasets total).}
				\label{table:training-samples}
			\end{sc}
		\end{tiny}
	\end{table}

	\section{Impact of LoRA Rank in Integration}\label{sec:rank}
	
	As described in Section~\ref{sec:training}, Stage~II integrates pretrained experts by training a shared linear router and applying LoRA adaptation to the attention layers. The router operates at the token level and selects one expert per token, while in standalone experts, the same adapter processes all tokens of a prompt. This change in routing granularity can alter token representations, so a small attention adaptation is introduced to improve the compatibility between token embeddings and routing decisions.
	
	In this section, we analyze whether Stage~II requires substantial parameter adaptation to learn expert selection. We vary the LoRA rank applied to the attention layers. The \emph{No LoRA} setting trains only the shared router while keeping the model representations unchanged.
	
	Table~\ref{table:ranks} reports the results when integrating the RLVF-trained experts from Stage~I. For both model sizes, allowing a small amount of attention adaptation improves performance. However, the gains quickly saturate. For Llama-8B, $R=32$ gives the highest score (79.93), while the \emph{No LoRA} setting already achieves 78.65, which is close to the maximum.
	
	These results indicate that Stage~II primarily learns routing behavior rather than relearning domain knowledge or reasoning ability. The attention LoRA stabilizes token-level routing, but only a small rank is sufficient, and larger ranks provide little additional benefit.
	
	\begin{table*}[!h]
		\centering
		\begin{small}
			\begin{sc}
				\begin{tabular}{l cccccc}
					\toprule
					\textbf{LoRA Rank} & GSM8K & ARC-C & MedQA & BoolQ & CoLA & Average \\
					\midrule
					\midrule
					\multicolumn{7}{l}{\texttt{Llama-3B}} \\
					\midrule
					\textbf{No Attention LoRA} & 63.46 & 70.82 & 53.62 & 81.44 & 68.19 & 67.51 \\
					\rowcolor{gray!5}
					\textbf{R}=16 & 65.81 & 75.26 & 51.61 & 82.17 & 72.29 & 69.43 \\
					\rowcolor{gray!10}
					\textbf{R}=32 & 70.74 & 76.79 & 52.63 & 83.30 & 75.36 & 71.76 \\
					\rowcolor{gray!15}
					\textbf{R}=64 & 72.10 & 76.79 & 52.32 & 83.55 & 75.55 & 72.06\\
					\rowcolor{gray!20}
					\textbf{R}=128 & 73.62 & 75.43 & 51.53 & 83.09 & 76.70 & \textbf{72.07} \\
					\midrule
					\midrule
					\multicolumn{7}{l}{\texttt{Llama-8B}} \\
					\midrule
					\textbf{No Attention LoRA} & 79.76 & 84.81 & 65.51 & 86.09 & 77.06 & 78.65 \\
					\rowcolor{gray!5}
					\textbf{R}=16 & 83.85 & 85.24 & 65.52 & 85.57 & 78.04 & 79.64 \\
					\rowcolor{gray!10}
					\textbf{R}=32 & 83.09 & 85.67 & 66.93 & 86.67 & 77.28 & \textbf{79.93} \\
					\rowcolor{gray!15}
					\textbf{R}=64 & 85.06 & 85.67 & 66.38 & 84.59 & 77.28 & 79.80 \\
					\rowcolor{gray!20}
					\textbf{R}=128 & 84.69 & 84.61 & 67.17 & 84.43 & 78.09 & 79.80 \\
					\bottomrule
				\end{tabular}
			\end{sc}
		\end{small}
		\caption{\textbf{Effect of attention LoRA rank during Stage~II integration using RLVF-trained experts.}  The configuration of \emph{No LoRA} trains only the shared router while keeping the backbone unchanged. Small ranks improve performance, but gains quickly saturate, indicating that integration mainly depends on routing rather than relearning expert knowledge.}
		\label{table:ranks}
	\end{table*}

    %%%%%%%%%%%%%%%%%%%%%%%%%%%%%%%%%%%%%%%%%%
	\section{Why Normalized Top-1 Soft Routing Has Zero Routing Gradient}
    \label{sec:top1-normalization-gradient}

    This section explains why a purely soft top-$1$ formulation cannot both preserve unit-scale LoRA application and provide a useful gradient for the router training.

    Let $p_i=\operatorname{softmax}(g(x))_i$ be the routing probability for expert $i$ while $g(x)$ is the logits produced by the router, and let $j=\arg\max_i p_i$ be the selected top-$1$ expert.

    A simple top-$1$ soft router can apply the selected expert as
    \begin{align}
        y_{\text{soft-top1}}(x)=p_j E_j(x)
    \end{align}
    This gives gradients to the router. For a fixed selected expert $j$,
    \begin{align}
        \frac{\partial y_{\text{soft-top1}}}{\partial g_\ell}
        =
        E_j(x)
        \frac{\partial p_j}{\partial g_\ell}
        =
        E_j(x)p_j(\mathbb{I}[\ell=j]-p_\ell)
    \end{align}
    However, this applies the selected LoRA expert with weight $p_j$ instead of weight $1$. Since $p_j < 1$ in general, the LoRA update is scaled down. This changes the unit-scale update that the expert used during standalone training.

    To keep the selected expert at unit scale, one may normalize the probability over the selected top-$1$ set $S=\{j\}$:
    \begin{align}
        \tilde p_j
        =
        \frac{p_j}{\sum_{m\in S}p_m}
        =
        \frac{p_j}{p_j}
        =
        1
    \end{align}
    The output then becomes
    \begin{align}
        y_{\text{norm-top1}}(x)
        =
        \tilde p_j E_j(x)
        =
        E_j(x)
    \end{align}
    This preserves the original LoRA scale. However, for a fixed selected expert $j$, the normalized weight is the constant function:
    \begin{align}
        \frac{\partial \tilde p_j}{\partial g_\ell}
        =
        \frac{\partial}{\partial g_\ell}
        \left(\frac{p_j}{p_j}\right)
        =
        0.
    \end{align}
    Thus, the normalized routing weight gives no useful gradient to the router.

    The only remaining dependence on the router is through the selection $j=\arg\max_i p_i$. This selection is discrete. It is piecewise constant and is not differentiable. Therefore, normalized top-$1$ routing preserves unit-scale expert application, but it cannot train the router with a useful task-loss
    gradient.

    Hard routing with a STE avoids this problem. In the forward pass, it selects one expert with unit scale:
    \begin{align}
        y_{\text{forward}}(x)
        =
        E_j(x).
    \end{align}
    In the backward pass, it uses the soft  probabilities as a surrogate gradient:
    \begin{align}
        Y
        =
        \operatorname{StopGradient}(Y_{\text{hard}} - p) + p.
    \end{align}
    This gives unit-scale expert selection in the forward pass while still allowing the router to be trained by gradient descent.

	%%%%%%%%%%%%%%%%%%%%%%%%%%%%%%%%%%%%%%%%%%
	
	\section{Results on Additional Model Families}\label{sec:additional}
	In addition to the LLaMA models used in the main experiments, we also evaluate our method on two other model families: Gemma-3 \cite{gemma3_technical_report_2025} and Qwen-$2.5$ \cite{qwen2025qwen25technicalreport}. The goal of this experiment is to verify that the observations reported in the paper are not limited to the LLaMA architecture.
	
	Tables~\ref{table:additional-fn-rlvf} and~\ref{table:additional-mixer} report the results for Gemma-3-4B and Qwen-$2.5$-7B, respectively. Table~\ref{table:additional-fn-rlvf} shows the standalone expert performance, while Table~\ref{table:additional-mixer} reports the integration results. As in the LLaMA experiments, RLVF produces stronger experts, and our method achieves the best integration performance while training far fewer parameters than LoRAMixer.
	
	Overall, these results show that the main findings of the paper remain consistent across different model families. This suggests that the proposed routing framework is general and can be applied to any instruction-tuned architectures, not only the LLaMA family.
	
	\begin{table*}[!h]
		\centering
		\begin{small}
			\begin{sc}
				\begin{tabular}{l  c c c c c c}
					\toprule
					Method & GSM8K & ARC-C & MedQA & BoolQ & CoLA & Avg \\
					\midrule
					\midrule
					\multicolumn{6}{l}{\texttt{Gemma-3-4B}} \\
					\midrule
					Baseline   & 85.43  & 82.62 & 44.79 & 80.98 & 80.63 & 74.89  \\
					\rowcolor{gray!5}
					SFT & 64.97 & 72.70 & 42.97 & \textbf{87.07} & 82.84 & 70.11 \\
					\rowcolor{gray!15}
					RLVF & \textbf{88.78} & \textbf{84.99} & \textbf{47.65} & 85.14 & \textbf{82.07} & \textbf{77.73} \\
					\midrule
					\midrule
					\multicolumn{6}{l}{\texttt{Qwen-$2.5$-$7$B}} \\
					\midrule
					Baseline & 54.13 & 89.07 & 57.42 & 81.41 & 83.03 & 73.01 \\
					\rowcolor{gray!5}
					SFT & 77.48 & 88.14 & \textbf{57.89} & 87.83 & \textbf{83.99} & 79.07 \\
					\rowcolor{gray!15}
					RLVF & \textbf{91.66} & \textbf{89.48} & 57.82 & \textbf{88.87} & 83.13 & \textbf{82.19} \\
					\bottomrule
				\end{tabular}
				\caption{
					\textbf{Performance of standalone experts on additional model families.}  We report results for Gemma-3-4B and Qwen-$2.5$-7B across five tasks.  As in the LLaMA experiments, RLVF improves reasoning performance compared to the baseline models.
				}
				\label{table:additional-fn-rlvf}
			\end{sc}
		\end{small}
	\end{table*}
	
	%%%%%%%%%%%%%%%%%%%%%%%%%%%%
	
	\begin{table*}[!h]
		\begin{center}
			\begin{small}
				\begin{sc}
					\begin{tabular}{l ccccccc}
						\toprule
						Method & GSM8K & ARC-C & MedQA & BoolQ & CoLA & Average &  \makecell{\# Trainable\\Parameters} \\
						\midrule
						\midrule
						\multicolumn{8}{l}{\texttt{Gemma-3-4B}} \\
						\midrule
						LoRAMixer TopK=1 & 86.05 & 83.70 & 46.50 & 80.94 & 80.44 & 75.53 & $\approx 836M$
						\\
						\rowcolor{gray!5}
						\makecell[l]{LoRAMixer TopK=2\\Normalized} & 86.88 & 83.10 & 47.68 & 83.43 & 81.30 & 76.48 & $\approx 836M$
						\\
						\rowcolor{gray!10}
						\textbf{Ours} & 87.57 & 83.61 & 48.78 & 84.80 & 81.59 & \textbf{77.27} & $\approx 72M$
						\\
						\midrule
						\midrule
						\multicolumn{8}{l}{\texttt{Qwen-$2.5$-$7$B}} \\
						\midrule
						LoRAMixer TopK=1 & 74.91 & 88.99 & 57.50 & 83.00 & 83.13 & 77.51 & $\approx 1.211B$
						\\
						\rowcolor{gray!5}
						\makecell[l]{LoRAMixer TopK=2\\Normalized} & 90.60 & 88.65 & 56.20 & 85.54 & 81.78 & 80.55 & $\approx 1.211B$
						\\
						\rowcolor{gray!10}
						\textbf{Ours} & 90.67 & 89.25 & 58.13 & 88.10 & 83.41 & \textbf{81.95} & $\approx 41M$
						\\
						\bottomrule
					\end{tabular}
					\caption{
						\textbf{Stage~II integration results on additional model families.}  We compare LoRAMixer and Hard-Routed MoR-LoRA when integrating pretrained experts.  For Qwen-$2.5$-7B, the mixer is trained with LoRA rank $r=64$.
					}
					\label{table:additional-mixer}
				\end{sc}
			\end{small}
		\end{center}
	\end{table*}

    \section{Comparison with Alternative Hard-Routing Approximations}\label{sec:hard-routing-approximations}

    Our main method uses deterministic hard top-$1$ routing with a straight-through estimator (STE).  This section compares it with alternative approximations for training a hard router while keeping all pretrained LoRA experts frozen. The goal is to determine whether the benefit comes from deterministic STE itself or from the broader property that the selected frozen LoRA expert is applied with unit scale during both training and inference.

    We compare three variants.

    \begin{enumerate}
        \item \textbf{Hard STE} is our default method: the forward pass selects the top-$1$ expert deterministically, while gradients are propagated through the soft router probabilities. 

        \item \textbf{Gumbel-Softmax} \cite{jang2016categorical} uses hard Gumbel-Softmax sampling with a continuous relaxation in the backward pass. 
        We anneal the temperature from $\tau=1.5$ to $\tau=0.1$ exponentially and keep the final temperature fixed for the last $10\%$ of training. 

        \item \textbf{Soft-train/hard-inference} trains the model with a probability-scaled top-$1$ expert output, but removes the softmax scaling at inference time and applies the selected expert with unit weight. 
        This variant tests whether unit-scale routing is sufficient only at inference time, or whether the same behavior is also necessary during training.
    \end{enumerate}

    All methods use the same Stage~II setting: RLVF-trained experts are frozen, the router is shared across layers, attention LoRA is enabled, and the model is trained on distilled traces with at most 1000 examples per dataset.

    Table~\ref{table:hard-routing-approximations} shows that Gumbel-Softmax performs similarly to deterministic STE but does not improve over it. On LLaMA-3B, Gumbel-Softmax reaches an average score of $71.82$, compared with $72.07$ for Hard STE. On LLaMA-8B, the gap is larger, with Gumbel-Softmax reaching $78.71$ compared with $79.80$ for Hard STE. This suggests that the exact surrogate-gradient estimator is not the main factor; both methods preserve the important forward-pass behavior of applying exactly one expert with unit scale.

    In contrast, soft-train/hard-inference performs substantially worse. The drop is especially large on LLaMA-3B, where the average score decreases from $72.07$ to $66.49$. This indicates that simply removing the softmax scaling at inference time is not sufficient. If the router and attention adaptation are trained under probability-scaled expert outputs, the model learns under one effective adapter function but is evaluated under another. Therefore, the unit-scale hard-routing behavior should be used consistently during both training and inference.

    \begin{table*}[!h]
        \centering
        \begin{small}
        \begin{sc}
        \begin{tabular}{l c c c c c c}
            \toprule
            Method & GSM8K & ARC-C & MedQA & BoolQ & CoLA & Avg \\
            \midrule
            \midrule
            \multicolumn{7}{l}{\texttt{LLaMA-3B}} \\
            \midrule
            \rowcolor{gray!15}
            \textbf{Hard STE} 
            & \textbf{73.62} & 75.43 & 51.53 & 83.09 & \textbf{76.70} & \textbf{72.07} \\
            Gumbel-Softmax 
            & 71.04 & \textbf{75.85} & \textbf{52.63} & \textbf{84.34} & 75.26 & 71.82 \\
            Soft-train/hard-inference 
            & 53.68 & 72.27 & 50.82 & 82.05 & 73.63 & 66.49 \\
            \midrule
            \midrule
            \multicolumn{7}{l}{\texttt{LLaMA-8B}} \\
            \midrule
            \rowcolor{gray!15}
            \textbf{Hard STE} 
            & \textbf{84.69} & \textbf{84.61} & \textbf{67.17} & \textbf{84.43} & 78.09 & \textbf{79.80} \\
            Gumbel-Softmax 
            & 83.55 & 84.47 & 65.04 & 83.88 & 76.61 & 78.71 \\
            Soft-train/hard-inference
            & 78.70 & 83.11 & 63.55 & 83.13 & \textbf{78.62} & 77.42 \\
            \bottomrule
        \end{tabular}
        \caption{
            \textbf{Comparison with alternative hard-routing approximations.}
            All methods use frozen RLVF-trained experts and train only the shared router and attention LoRA during Stage~II. 
            Gumbel-Softmax is competitive but does not improve over deterministic STE. 
            Soft-train/hard-inference performs worse, showing that unit-scale hard routing should be used during training as well as inference.
        }
        \label{table:hard-routing-approximations}
        \end{sc}
        \end{small}
    \end{table*}

    We further compare STE and Gumbel-Softmax under different Stage~II data budgets on LLaMA-3B. This analysis tests whether the routing approximation remains stable when the amount of routing supervision is reduced.

    As shown in Table~\ref{table:gumbel-data-budget}, Gumbel-Softmax is competitive when 1000 samples per dataset are available, trailing deterministic STE by only $0.25$ average points. However, when the budget is reduced to 750 samples per dataset, the gap increases to $2.47$ average points. Because the pretrained experts are frozen, stochastic expert sampling cannot improve the experts themselves; it only changes which fixed expert is selected during router learning. This additional routing variance can be harmful when Stage~II supervision is limited.
    
    \begin{table*}[!h]
        \centering
        \begin{small}
        \begin{sc}
        \begin{tabular}{l c c c c c c}
            \toprule
            Method & GSM8K & ARC-C & MedQA & BoolQ & CoLA & Avg \\
            \midrule
            \midrule
            \multicolumn{7}{l}{\texttt{1000 Samples}} \\
            \midrule
            Hard STE
            & \textbf{73.62} & 75.43 & 51.53 & 83.09 & \textbf{76.70} & \textbf{72.07} \\
            Gumbel-Softmax 
            & 71.04 & \textbf{75.85} & \textbf{52.63} & \textbf{84.34} & 75.26 & 71.82 \\
            \midrule
            \midrule
            \multicolumn{7}{l}{\texttt{750 Samples}} \\
            \midrule
            Hard STE       
            & \textbf{70.36} & \textbf{73.72} & 52.57 & \textbf{79.88} & \textbf{74.45} & \textbf{70.20} \\
            Gumbel-Softmax 
            & 64.14 & 72.35 & \textbf{52.63} & 79.33 & 70.18 & 67.73 \\
            \bottomrule
        \end{tabular}
        \caption{
            \textbf{Effect of Stage~II data budget on deterministic STE and Gumbel-Softmax routing for LLaMA-3B.}
            With 1000 samples per dataset, Gumbel-Softmax is close to deterministic STE. 
            With 750 samples per dataset, the gap increases, suggesting that stochastic routing is less stable when routing supervision is limited.
        }
        \label{table:gumbel-data-budget}
        \end{sc}
        \end{small}
    \end{table*}

    \section{Reproducibility Details}\label{sec:reproducibility}

    This section provides additional implementation and evaluation details for reproducing the experiments. All datasets used in this work are public benchmarks. Unless otherwise stated, the same training configuration is used across all model families, including LLaMA, Gemma, and Qwen; only the underlying instruction-tuned backbone model changes. Additionally, the code is available at \href{https://github.com/sar-molavi/hard-routed-mor-lora}{github.com/sar-molavi/hard-routed-mor-lora}.

    \subsection{Datasets and Evaluation Protocol}
    The main experiments use five datasets, corresponding to five domains: mathematical reasoning, commonsense reasoning, medical question answering, reading comprehension, and grammatical acceptability. Specifically, we use GSM8K \cite{cobbe2021training}, ARC-Challenge \cite{clark2018think}, MedQA \cite{jin2021disease}, BoolQ \cite{clark2019boolq}, and CoLA \cite{warstadt2019neural}. Additional unseen-dataset evaluation uses SVAMP \cite{patel2021nlp} and SST-2 \cite{socher2013recursive}. SVAMP and SST-2 are not used for Stage~I expert training or Stage~II mixer training.

    All reported results are evaluated with greedy decoding. The evaluation metric is accuracy. Table~\ref{table:dataset-reproducibility} reports the public dataset sources and the number of evaluation examples used for each benchmark.

    \begin{table*}[!h]
        \centering
        \begin{small}
        \begin{tabular}{l l l r}
            \toprule
            \textbf{Dataset} & \textbf{Use} & \textbf{Source} & \textbf{\# Eval. Samples} \\
            \midrule
            GSM8K & Main evaluation & \url{https://huggingface.co/datasets/openai/gsm8k} & 1,319 \\
            ARC-Challenge & Main evaluation & \url{https://huggingface.co/datasets/allenai/ai2_arc} & 1,172 \\
            MedQA & Main evaluation & \url{https://huggingface.co/datasets/bigbio/med_qa} & 1,273 \\
            BoolQ & Main evaluation & \url{https://huggingface.co/datasets/google/boolq} & 3,270 \\
            CoLA & Main evaluation & \url{https://huggingface.co/datasets/nyu-mll/glue} & 1,043 \\
            SVAMP & Unseen-dataset evaluation & \url{https://huggingface.co/datasets/ChilleD/SVAMP} & 300 \\
            SST-2 & Unseen-dataset evaluation & \url{https://huggingface.co/datasets/nyu-mll/glue} & 872 \\
            \bottomrule
        \end{tabular}
        \end{small}
        \caption{
            \textbf{Datasets and evaluation split sizes.}
            The main five-domain experiments use GSM8K, ARC-Challenge, MedQA, BoolQ, and CoLA. SVAMP and SST-2 are used only for the unseen-dataset evaluation.
        }
        \label{table:dataset-reproducibility}
    \end{table*}

    \subsection{Stage~I: Standalone Expert Training}
    In Stage~I, we train one LoRA expert independently for each dataset. The base model is frozen, and only the LoRA parameters are updated. We train two types of standalone experts: supervised fine-tuning (SFT) experts and reinforcement learning from verifiable feedback (RLVF) experts.

    For both SFT and RLVF experts, LoRA is applied to the feed-forward network projections: \texttt{gate\_proj}, \texttt{up\_proj}, and \texttt{down\_proj}.
    The SFT experts are trained with supervised next-token likelihood. The RLVF experts are trained with the off-policy GRPO-style objective described in Section~\ref{sec:RL}, using automatically verifiable rewards computed from final answers. The full Stage~I training configuration, including LoRA target modules, rank, learning rate, batch size, gradient accumulation, number of epochs, precision, and maximum sequence length, is reported in Table~\ref{table:training-reproducibility}.

    \subsection{RLVF Reward and Verifier Construction}
    \label{sec:rlvf-reward}

    RLVF uses automatically verifiable rewards computed from final answers. We do not use human preference labels. For each prompt, multiple completions are sampled from a behavior policy, which is a few checkpoints behind the target policy. The target policy is then optimized using a token-level off-policy GRPO-style objective with importance ratios between the target policy and the behavior policy as detailed in Section~\ref{sec:RL}.

    The reward is $1.0$ for a correct answer and $-1.1$ for an incorrect answer. We also use two small formatting bonuses: a think-format bonus of $0.2$ when the response follows the expected explicit thinking format, and a JSON-on-wrong bonus of $0.1$ when an incorrect response still produces a valid structured JSON answer. These bonuses are used to stabilize response formatting and answer extraction during RLVF training. The expected response format consists of an explicit reasoning block followed by a structured JSON answer:
    \begin{center}
        \verb|<think>...<\think>|
        \verb|{"answer": <...>}|
    \end{center}

    Table~\ref{table:rlvf-reproducibility} summarizes the RLVF configuration.
    
    \begin{table}[!h]
        \centering
        \begin{small}
        \begin{tabular}{l c}
            \toprule
            \textbf{RLVF Setting} & \textbf{Value} \\
            \midrule
            Objective & Offline Token-level GRPO \\
            Advantage mode & mean-reward \\
            Ratio clipping & 0.2 \\
            Generations per prompt & 4 \\
            Sampling temperature & 1.1 \\
            Max new tokens & 3072 \\
            checkpoint refresh interval & 20 training steps \\
            Correct / incorrect reward & $1.0 / -1.1$ \\
            Think-format bonus & 0.2 \\
            JSON-on-wrong bonus & 0.1 \\
            \bottomrule
        \end{tabular}
        \end{small}
        \caption{
            \textbf{RLVF configuration.}
            Rewards are computed using automatically verifiable final answers. The formatting bonuses are used only to stabilize reasoning and answer extraction format.
        }
        \label{table:rlvf-reproducibility}
    \end{table}

    \subsection{Stage~II: Distillation and Mixer Training}
    Stage~II integrates the pretrained experts into one shared model. All Stage~I experts remain frozen. We train only a shared linear router and a lightweight LoRA adaptation on the attention projections.

    To construct the Stage~II training data, shown in Figure~\ref{fig:pipeline}, we query each frozen expert using prompts from its own domain and collect the generated reasoning traces. These traces form the supervised distillation dataset. The mixer is then trained with standard next-token likelihood on the distilled traces. The objective of Stage~II is to learn expert selection rather than to relearn the domain knowledge or reasoning ability encoded in the frozen experts.

    Unless otherwise stated, Stage~II uses at most 1000 prompts per dataset. The attention LoRA is applied to: \texttt{q\_proj}, \texttt{k\_proj}, \texttt{v\_proj}, and \texttt{o\_proj} with a low rank.
    The router is shared across layers, and routing is performed at the token level. During training and inference, the selected LoRA expert is applied with unit scale.

    \subsection{Training Hyperparameters}
    Table~\ref{table:training-reproducibility} summarizes the main training configuration for Stage~I and Stage~II. The same configuration is used across LLaMA, Gemma, and Qwen experiments unless explicitly stated otherwise. The global batch size is $32$ in all stages, and we use gradient accumulation and gradient checkpointing.

    \begin{table*}[!h]
		\centering
		\renewcommand{\arraystretch}{1.28}
		\setlength{\tabcolsep}{5pt}
		\newcolumntype{Y}{>{\raggedright\arraybackslash}X}
		\arrayrulecolor{gray!35}

		\begin{small}

		\textbf{Model and Trainable Components}
		\vspace{0.35em}

		\begin{tabularx}{\textwidth}{>{\bfseries\raggedright\arraybackslash}p{0.23\textwidth}YYY}
			\toprule
			\rowcolor{gray!15}
			\textbf{Setting}
			& \textbf{Stage~I: SFT Experts}
			& \textbf{Stage~I: RLVF Experts}
			& \textbf{Stage~II: Mixer} \\
			\midrule
			\rowcolors{2}{gray!6}{white}

			Base model
			& Experiment-specific backbone
			& Experiment-specific backbone
			& Experiment-specific backbone \\
			\hline

			Trainable modules
			& FFN LoRA
			& FFN LoRA
			& Shared router + attention LoRA \\
			\hline

			Frozen modules
			& Backbone
			& Backbone
			& Backbone + expert LoRAs \\
			\hline

			LoRA target modules
			& \texttt{gate\_proj}, \texttt{up\_proj}, \texttt{down\_proj}
			& \texttt{gate\_proj}, \texttt{up\_proj}, \texttt{down\_proj}
			& \texttt{q\_proj}, \texttt{k\_proj}, \texttt{v\_proj}, \texttt{o\_proj} \\
			\hline

			LoRA rank / alpha / dropout
			& $128 / 256 / 0.1$
			& $128 / 256 / 0.1$
			& $128 / 256 / 0.1$ \\
			\hline

			Objective
			& Next-token likelihood
			& Offline GRPO-style RLVF
			& Next-token likelihood on distilled traces \\
			\bottomrule
		\end{tabularx}

		\vspace{1.0em}

		\textbf{Optimization and Sequence Settings}
		\vspace{0.35em}

		\begin{tabularx}{\textwidth}{>{\bfseries\raggedright\arraybackslash}p{0.23\textwidth}YYY}
			\toprule
			\rowcolor{gray!15}
			\textbf{Setting}
			& \textbf{Stage~I: SFT Experts}
			& \textbf{Stage~I: RLVF Experts}
			& \textbf{Stage~II: Mixer} \\
			\midrule
			\rowcolors{2}{gray!6}{white}

			Learning rate
			& $1\mathrm{e}{-6}$
			& $5\mathrm{e}{-6}$
			& $1\mathrm{e}{-6}$ \\
			\hline

			Epochs
			& 10
			& 2
			& 1 \\
			\hline

			Batch size
			& 32
			& 32
			& 32 \\
			\hline

			LR scheduler / warmup
			& Cosine / 0.05
			& Cosine / 0.05
			& Cosine / 0.05 \\
			\hline

			Weight decay
			& 0.1
			& 0.0
			& 0.0 \\
			\hline

			Max grad norm
			& 1.0
			& 0.75
			& 1.0 \\
			\hline

			Precision
			& bf16
			& bf16
			& bf16 \\
			\hline

			Max sequence length
			& No explicit limit
			& 5120
			& 5120 \\
			\hline

			Stage~II samples per dataset
			& --
			& --
			& 1000 \\
			\bottomrule
		\end{tabularx}

		\end{small}

		\arrayrulecolor{black}

		\caption{
			\textbf{Training configuration for Stage~I and Stage~II.}
			Stage~I trains standalone dataset-specific experts while keeping the backbone frozen. Stage~II freezes all pretrained experts and trains only the shared router and attention LoRA. The same configuration is used across model families unless otherwise stated.
		}
		\label{table:training-reproducibility}
	\end{table*}

    \subsection{RLVF Training Dynamics}
    To make the behavior of Stage~I RLVF training more transparent, we report training dynamics for the standalone RLVF experts. We track three quantities during training: completion length, verifier reward, and stop rate, whether the generation hit the end-of-sentence token. These statistics are collected from sampled trajectories during RLVF optimization and are, therefore, different from the final benchmark results, which are evaluated separately using greedy decoding.

    For each training step, we compute the per-step mean over the few sampled completions. We also report a bias-corrected exponential moving average (EMA) with decay $0.95$ to show the overall trend more clearly. The curves cover the five main domains, \textsc{ARC-C}, \textsc{BoolQ}, \textsc{CoLA}, \textsc{GSM8K}, and \textsc{MedQA}, for both LLaMA-3B and LLaMA-8B experts.
    
    The reward is computed using the same automatic verifier used during RLVF training. Since trajectories are sampled with high-temperature exploration on a few training samples using offline reinforcement learning, the reward curves are expected to fluctuate and should not be interpreted as final task accuracy.

\begin{figure*}[!h]
        \centering
        \includegraphics[width=0.95\linewidth]{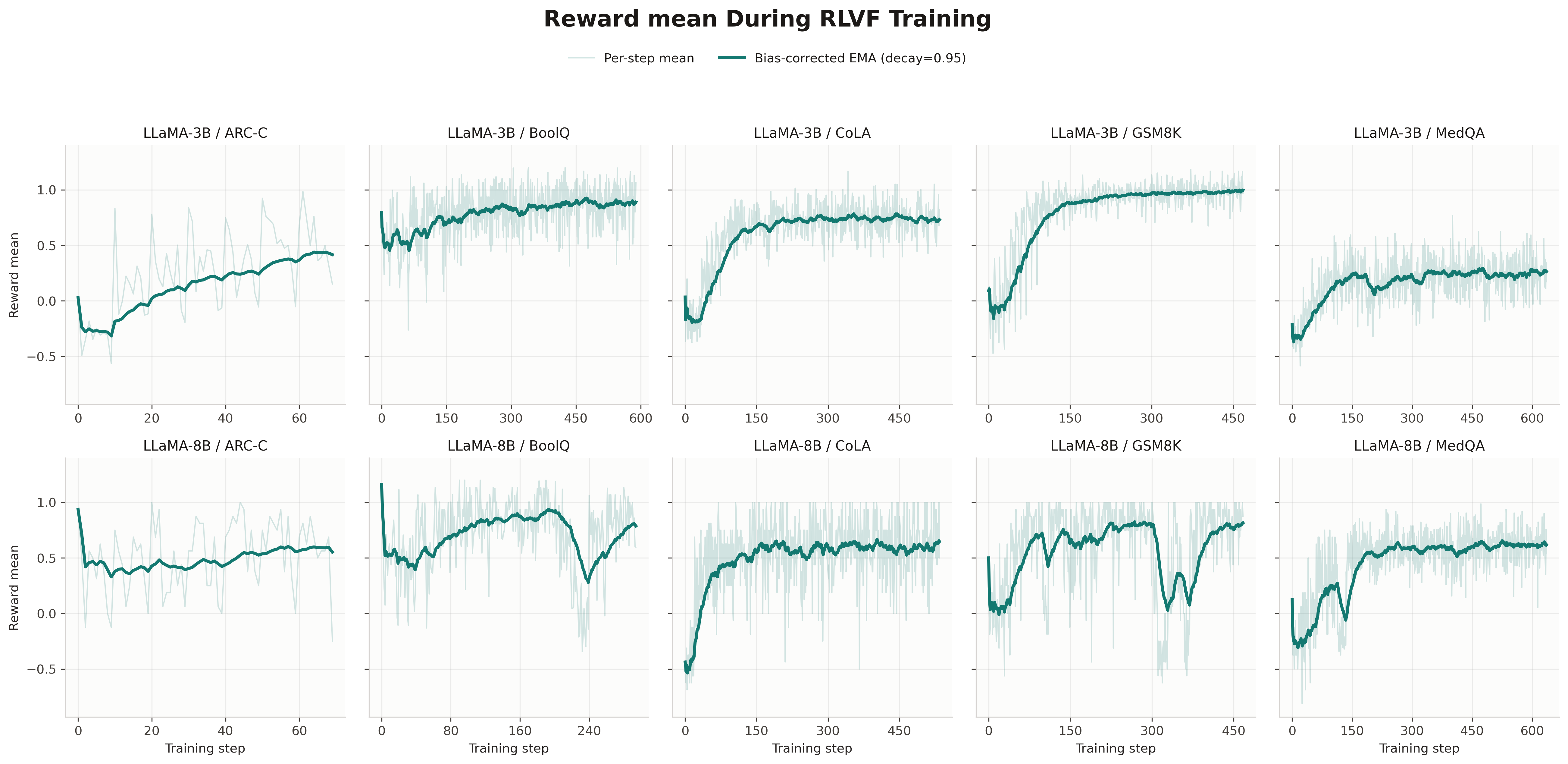}
        \caption{
            \textbf{Reward mean during RLVF training.}
            We report the mean verifiable reward obtained by the sampled trajectories during RLVF training. Rows correspond to model scales and columns correspond to datasets. The light curves show the per-step mean, while the darker curves show a bias-corrected exponential moving average with decay $0.95$. Across most datasets, the smoothed reward increases during training, indicating that the experts progressively improve with respect to the automatic verifier. The fluctuations are expected because trajectories are sampled with high-temperature exploration on a few samples. These are training reward statistics, not final benchmark accuracies.
        }
        \label{fig:rlvf-reward-mean}
    \end{figure*}

    Figure~\ref{fig:rlvf-reward-mean} reports the mean verifier reward. Across most datasets and model sizes, the smoothed reward increases during training. This indicates that RLVF improves the sampled trajectories with respect to automatically verifiable correctness. The trend is especially clear for datasets such as \textsc{GSM8K}, \textsc{BoolQ}, and \textsc{CoLA}, where the reward rises steadily after the early training phase. Some non-monotonic behavior remains, particularly for the larger model on certain tasks, which is expected because the policy continues to sample diverse trajectories during training. Overall, the reward dynamics support the main Stage~I observation that RLVF improves standalone experts by optimizing verifiable correctness rather than imitating fixed reference traces.
    
\begin{figure*}[!h]
        \centering
        \includegraphics[width=0.95\linewidth]{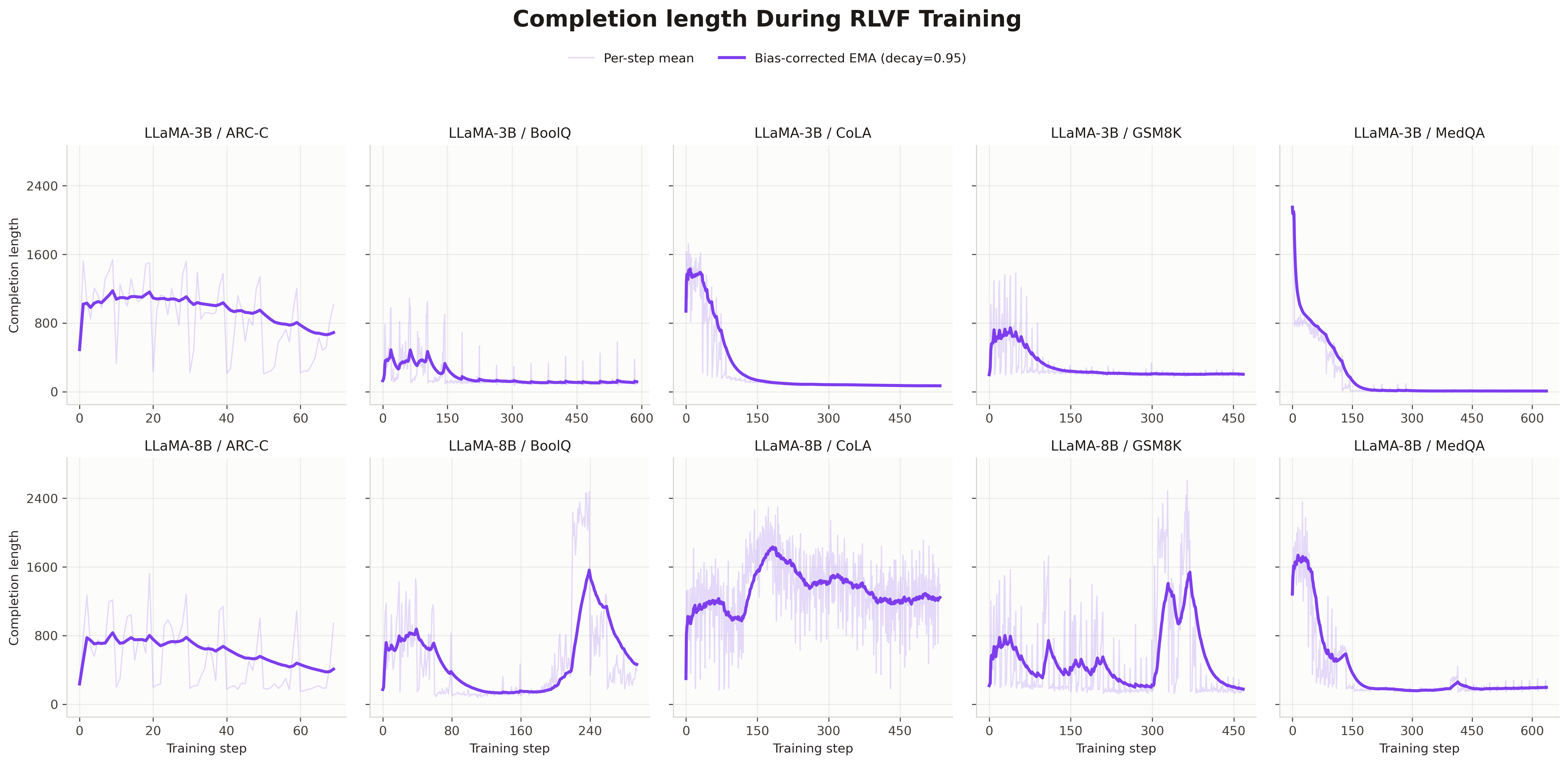}
        \caption{
            \textbf{Completion length during RLVF training.}
            We report the mean generated completion length for each RLVF-trained expert across training steps. Rows correspond to model scales and columns correspond to datasets. The light curves show the per-step mean, while the darker curves show a bias-corrected exponential moving average with decay $0.95$. Completion length varies substantially across datasets and model sizes, reflecting differences in task format, reasoning behavior, and exploration under high-temperature sampling. These curves are used only to characterize training dynamics and should not be interpreted as the final benchmark performance.
        }
        \label{fig:rlvf-completion-length}
    \end{figure*}

    Figure~\ref{fig:rlvf-completion-length} shows the generated completion length during training. The absolute completion length differs substantially across datasets because the tasks require different output formats and reasoning depths. For several experts, completion length decreases or stabilizes as training progresses, suggesting that the model learns to produce more concise outputs that satisfy the verifier. However, completion length itself is not optimized as the primary objective, and longer or shorter generations should not be interpreted as better performance without considering the corresponding reward and stopping behavior.

\begin{figure*}[!h]
        \centering
        \includegraphics[width=0.95\linewidth]{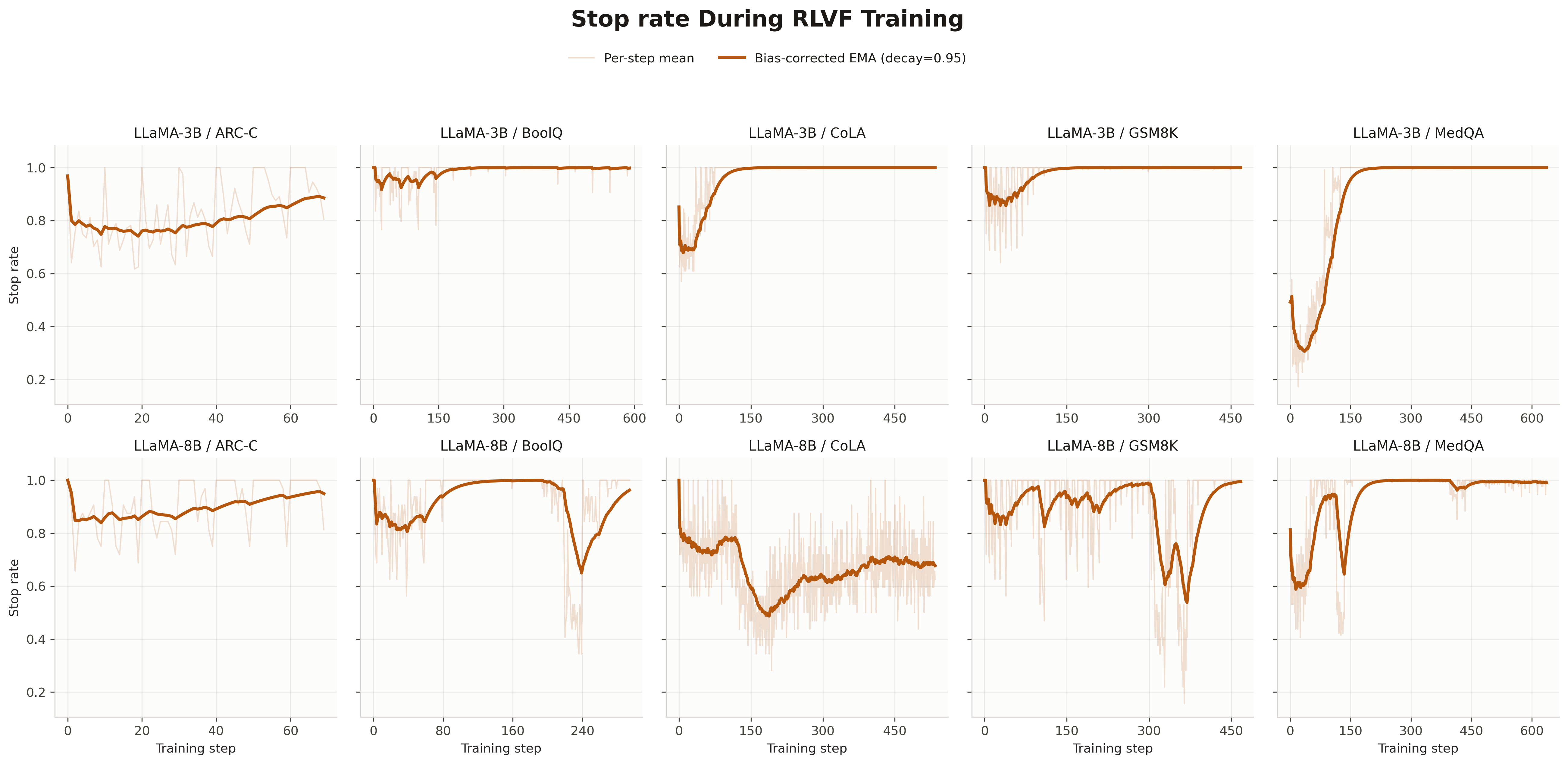}
        \caption{
            \textbf{Stop rate during RLVF training.}
            We report the fraction of sampled completions that satisfy the stopping criterion during RLVF training. Rows correspond to model scales and columns correspond to datasets. The light curves show the per-step stop rate, while the darker curves show a bias-corrected exponential moving average with decay $0.95$. For most experts, the stop rate increases or remains high as training progresses, suggesting that the models learn to produce complete outputs under the required generation format. Temporary drops and fluctuations reflect continued exploration during sampling and should be interpreted together with the reward and completion-length curves.
        }
        \label{fig:rlvf-stop-rate}
    \end{figure*}

    Finally, Figure~\ref{fig:rlvf-stop-rate} shows the stop rate, defined as the fraction of sampled completions that satisfy the generation stopping criterion before reaching the maximum generation budget. This statistic helps distinguish useful changes in completion length from unstable generation behavior. For most experts, the stop rate either remains high or increases toward a stable value during training, indicating that the models generally learn to terminate their responses correctly under the required output format. Some datasets exhibit temporary drops or more variable stopping behavior, especially under continued high-temperature exploration. These fluctuations are consistent with the reward dynamics and do not necessarily indicate degradation in final greedy-decoding performance.

    \section{Prompt-Level Routing and Additional Evaluation}
    \label{sec:prompt-routing-details}

    This appendix provides additional details for the prompt-level routing baseline, the mixed-domain evaluation, and the unseen-dataset evaluation discussed in the paper.

    \subsection{Prompt-Level Routing Baseline}

    The prompt-level routing baseline selects one expert for the entire input. We train a \texttt{bert-base-uncased} classifier to predict the domain of the input prompt. The predicted domain is then used to select the corresponding frozen LoRA expert for the whole generation. This baseline uses the same Stage~II supervision budget as our mixer: at most 1000 examples per dataset.

    The classifier achieves near-perfect domain classification accuracy, as shown in Table~\ref{table:prompt-classifier-accuracy}. This confirms that the original single-domain evaluation contains a strong domain-classification component.

    Table~\ref{table:prompt-routing-full} reports the full per-task results for prompt-level routing on LLaMA-3B and LLaMA-8B with RLVF-trained experts. Prompt-level routing slightly outperforms Hard-Routed MoR-LoRA on average in the clean single-domain setting. This supports the conclusion in the main paper: when each prompt clearly belongs to one domain, selecting one expert for the whole prompt is a strong and simple solution.

    \begin{table}[!h]
        \centering
        \begin{small}
        \begin{sc}
        \begin{tabular}{l c}
            \toprule
            Component & Value \\
            \midrule
            Prompt-level classifier & \texttt{bert-base-uncased} \\
            Classifier parameters & $\approx$ 110M \\
            Max train samples & 1000 \\
            Number of domains & 5 \\
            Training epochs & 3 \\
            Learning rate & 2e-5 \\
            Weight decay & 0.01 \\
            Warmup ratio & 0.06 \\
            Batch size & 32 \\
            Max sequence length & 256 \\
            \bottomrule
        \end{tabular}
        \end{sc}
        \end{small}
        \caption{
        \textbf{Prompt-level classifier setup.}
        The classifier predicts the input domain and selects one frozen LoRA expert for the whole prompt.
        }
        \label{table:prompt-classifier-setup}
    \end{table}

    \begin{table}[!h]
        \centering
        \begin{small}
        \begin{sc}
        \begin{tabular}{l c}
            \toprule
            Dataset & Accuracy \\
            \midrule
            GSM8K & 100.00 \\
            ARC-C & 98.89 \\
            MedQA & 99.61 \\
            BoolQ & 100.00 \\
            CoLA & 99.42 \\
            \midrule
            Overall & 99.70 \\
            \bottomrule
        \end{tabular}
        \end{sc}
        \end{small}
        \caption{
        \textbf{Domain classification accuracy of the prompt-level classifier.}
        The high accuracy explains why prompt-level routing is a strong baseline on clean single-domain inputs.
        }
        \label{table:prompt-classifier-accuracy}
    \end{table}

    \begin{table*}[!h]
        \centering
        \begin{small}
        \begin{sc}
        \begin{tabular}{l cccccc}
            \toprule
            Method & GSM8K & ARC-C & MedQA & BoolQ & CoLA & Avg \\
            \midrule
            \midrule
            \multicolumn{7}{l}{\texttt{LLaMA-3B}} \\
            \midrule
            Prompt-level routing & \textbf{75.36} & \textbf{76.02} & 51.13 & \textbf{85.33} & 76.13 & \textbf{72.79} \\
            \rowcolor{gray!10}
            Hard-Routed MoR-LoRA & 73.62 & 75.43 & \textbf{51.53} & 83.09 & \textbf{76.70} & 72.07 \\
            \midrule
            \midrule
            \multicolumn{7}{l}{\texttt{LLaMA-8B}} \\
            \midrule
            Prompt-level routing & 83.05 & 82.84 & 66.71 & \textbf{87.67} & \textbf{79.09} & \textbf{79.87}\\
            \rowcolor{gray!10}
            Hard-Routed MoR-LoRA & \textbf{84.69} & \textbf{84.61} & \textbf{67.17} & 84.43 & 78.09 & 79.80 \\
            \bottomrule
        \end{tabular}
        \end{sc}
        \end{small}
        \caption{
        \textbf{Full prompt-level routing results on clean single-domain inputs.}
        Prompt-level routing selects one frozen expert for the whole input, while Hard-Routed MoR-LoRA routes at the token level.
        }
        \label{table:prompt-routing-full}
    \end{table*}

    \subsection{Mixed-Domain Evaluation Details}

    The prompt-level baseline is strong when each input belongs to a single domain, but it must select one expert for the entire prompt. To test this limitation, we construct a mixed-domain evaluation using GSM8K and BoolQ. Each input contains one GSM8K math problem and one BoolQ question, and the model is asked to produce two structured answers: \texttt{math\_answer} and \texttt{boolq\_answer}. No method is trained on mixed-domain prompts. Table~\ref{table:mixed-domain-setup} details the process.

    We choose GSM8K and BoolQ because the prompt-level classifier reaches 100.00\% accuracy on both datasets in the original single-domain setting. Thus, in the clean setting, the prompt-level classifier can reliably identify both domains. The mixed-domain setting tests a different issue: whether selecting only one expert for the whole prompt is sufficient when the input contains two different types of questions.

    \begin{table*}[!h]
        \centering
        \begin{small}
        \begin{sc}
        \begin{tabular}{l p{0.58\linewidth}}
            \toprule
            Field & Description \\
            \midrule
            Mixed domains & GSM8K + BoolQ \\
            Prompt content & One math problem and one BoolQ question \\
            Expected output & Two structured answers: \texttt{math\_answer} and \texttt{boolq\_answer} \\
            Prompt order & Randomly sampled from \texttt{math\_first} and \texttt{boolq\_first} \\
            Training exposure & No method was trained on mixed-domain prompts \\
            Reason for choosing these domains & Both have 100.00\% prompt-level classification accuracy in the original single-domain setting \\
            \bottomrule
        \end{tabular}
        \end{sc}
        \end{small}
        \caption{
        \textbf{Mixed-domain evaluation setup.}
        The setting tests whether one expert is sufficient when a single input contains questions from two different domains.
        }
        \label{table:mixed-domain-setup}
    \end{table*}

    Prompt-level routing performs better on the math part, while Hard-Routed MoR-LoRA performs substantially better on the BoolQ part and obtains higher average performance. This shows the structural limitation of prompt-level routing: it can only choose either the GSM8K expert or the BoolQ expert for the whole input. In contrast, token-level routing is not restricted to one expert for the entire prompt. Furthermore, the token-level routing can be optimized for this situation, contrary to the prompt-level setting.

    \subsection{Unseen-Dataset Evaluation}\label{sec:ood-eval}

    We further evaluate whether the learned router and frozen experts transfer to related datasets that are not used during Stage~I expert training or Stage~II mixer training. We use SVAMP for mathematical reasoning and SST-2 for sentiment classification. Here, Baseline denotes the original instruction-tuned model without LoRA experts or routing. This experiment tests transfer to unseen but related task distributions.

    As shown in Table~\ref{table:ood-eval}, Hard-Routed MoR-LoRA improves over the original instruction-tuned model on both unseen datasets and both model scales. On SVAMP, the routed model transfers arithmetic reasoning behavior to a math dataset that was not used during training. On SST-2, which is also not one of the original five domains, the routed model improves over the baseline as well. These results suggest that the router is not limited to memorizing the original training datasets, and that frozen expert behavior can transfer to related unseen inputs. However, this experiment should be interpreted as transfer to related task distributions rather than general open-domain robustness.
    
    \begin{table}[!h]
        \centering
        \begin{small}
        \begin{sc}
        \begin{tabular}{lccc}
            \toprule
            Method & SST-2 & SVAMP & Avg \\
            \midrule
            \midrule
            \multicolumn{4}{l}{\texttt{LLaMA-3B}} \\
            \midrule
            Baseline & 75.11 & 1.67 & 38.39 \\
            \rowcolor{gray!10}
            Ours & \textbf{88.76} & \textbf{81.33} & \textbf{85.05} \\
            \midrule
            \midrule
            \multicolumn{4}{l}{\texttt{LLaMA-8B}} \\
            \midrule
            Baseline & 82.68 & 82.00 & 82.34 \\
            \rowcolor{gray!10}
            Ours & \textbf{90.71} & \textbf{85.67} & \textbf{88.19} \\
            \bottomrule
        \end{tabular}
        \end{sc}
        \end{small}
        \caption{
        \textbf{Evaluation on unseen datasets.}
        SVAMP and SST-2 are not used during expert training or mixer training. The results evaluate transfer to related unseen task distributions.
        }
        \label{table:ood-eval}
    \end{table}
    
\end{document}